\documentclass[11pt]{article}

\usepackage[preprint]{acl}

\usepackage{times}
\usepackage{latexsym}

\usepackage[T1]{fontenc}

\usepackage[utf8]{inputenc}

\usepackage{microtype}

\usepackage{inconsolata}

\usepackage{graphicx}

\usepackage{hyperref}
\usepackage{url}
\usepackage{algorithm}
\usepackage{algorithmicx}
\usepackage{algpseudocode}
\usepackage{booktabs}
\usepackage{multirow}
\usepackage[table]{xcolor}
\usepackage{cleveref}
\usepackage{wrapfig}
\usepackage{float}
\usepackage{amssymb}

\title{\fontsize{18}{22}\selectfont \textbf{Efficient Self-Evaluation for Diffusion Language Models via Sequence Regeneration}}

\author{
\begin{tabular}{l}
Linhao Zhong$^{1}$\thanks{Equal Contribution.} \quad Linyu Wu$^{2}$\footnotemark[1] \quad Wen Wang$^{1}$ \quad Yuling Xi$^{1}$ \quad Chenchen Jing$^{1,3}$ \\ Jiaheng Zhang$^{2}$ \quad Hao Chen$^{1}$  \quad Chunhua Shen$^{1,3}$\thanks{Corresponding Author.} \\
\textnormal{$^{1}$Zhejiang University} \quad
\textnormal{$^{2}$National University of Singapore} \quad
\textnormal{$^{3}$Zhejiang University of Technology}
\end{tabular}
}

\begin{document}
\maketitle
\begin{abstract}

Diffusion large language models (dLLMs) have recently attracted significant attention for their ability to enhance diversity, controllability, and parallelism.
However, their non-sequential, bidirectionally masked generation makes quality assessment difficult, underscoring the need for effective self-evaluation. 
In this work, we propose DiSE, a simple yet effective self-evaluation confidence quantification method for dLLMs. 
DiSE quantifies confidence by computing the probability of regenerating the tokens in the entire generated sequence, given the full context. 
This method enables more efficient and reliable quality assessment by leveraging token regeneration probabilities, facilitating both likelihood estimation and robust uncertainty quantification. 
Building upon DiSE, we further introduce a flexible-length generation framework, which adaptively controls the sequence length based on the model's self-assessment of its own output.
We analyze and validate the feasibility of DiSE from the perspective of dLLM generalization, and empirically demonstrate that DiSE is positively correlated with both semantic coherence and answer accuracy. 
Extensive experiments on likelihood evaluation, uncertainty quantification, and flexible-length generation further confirm the effectiveness of the proposed DiSE.
\textbf{Project page:} \url{https://zhongzero.github.io/DiSE}

\end{abstract}
\section{Introduction}

Recently, diffusion large language models (dLLMs)~\citep{llada, llada1.5, dream2025,dLLM-survey} have emerged as a promising direction in natural language processing. 
In contrast to auto-regressive (AR) models, dLLMs adopt the generative framework of diffusion models~\citep{DDPM, IDDPM, DDIM}, framing text generation as a progressive denoising process. 
This approach enables better diversity, controllability, and parallel generation compared to AR models. 
Nonetheless, the non-sequential and bidirectional nature of dLLMs makes direct likelihood-based self-evaluation challenging~\citep{llada}.
Concurrently, self-evaluation has been recognized as a fundamental capability of LLMs, serving as the basis for a wide range of applications such as hallucination detection~\citep{uq-survey1, fadeeva2024fact}, answer quality assessment~\citep{chang2024survey}, and generation quality enhancement~\citep{huang2024calibrating, xie2024calibrating}.

In AR models, causal masking enforces a strict left-to-right generation order, allowing sequence probability to be decomposed into token-level conditional probabilities.
This simplifies the generation process and enables self-evaluation through likelihood estimation. 
In contrast, dLLMs use bidirectional masking and a non-sequential, stepwise generation process, making direct likelihood-based self-evaluation challenging.
Currently, dLLMs rely primarily on Monte Carlo simulation-based approximations of sequence likelihood~\citep{llada}, but this method is computationally expensive and often yields suboptimal estimates, limiting its practical effectiveness. 
Moreover, owing to the intrinsic token-level self-evaluation signal provided by next-token prediction in AR models, the generation length can be adaptively controlled via real-time EOS token prediction.
Unlike AR models, conventional dLLMs lack such an effective built-in likelihood-based self-evaluation signal, which forces them into fixed-length generation and fundamentally restricts their flexibility.

In this paper, we aim to explore a more efficient and effective self-evaluation technique and its applications for dLLMs by addressing the following two research questions (RQ):
\begin{itemize}
\item \textbf{RQ1.} \textit{How can likelihood-based self-evaluation for dLLMs be made both faster and more effective, and how can such a method be made interpretable and empirically verified?}
\item \textbf{RQ2.} \textit{What are the practical benefits of obtaining a more efficient self-evaluation method for dLLMs?}
\end{itemize}

\noindent \textbf{Contribution 1.} 
To answer the first question, we propose DiSE, a simple yet effective self-evaluation confidence quantification method for diffusion large language models. 
DiSE is derived by feeding the entire sequence back into the dLLM and computing the probability of regenerating its tokens under the full context. 
This method enables the model to assess its own generation quality by evaluating how well it can reproduce the original sequence when conditioned on the entire context, effectively leveraging its own internal predictions. 
From an interpretability perspective, we analyze DiSE in terms of the dLLM's generalization capabilities. We explain and empirically validate the robustness of dLLMs against input perturbations and the feasibility of using token regeneration as a confidence measure.  
From an empirical perspective, we demonstrate through experiments that DiSE is positively correlated with both semantic coherence and answer accuracy.

\noindent \textbf{Contribution 2.} To answer the second question, we apply DiSE to three different aspects. 
DiSE provides a versatile mechanism for dLLMs, acting as an effective estimator for conditional likelihood evaluation and facilitating robust uncertainty quantification~\citep{uq-survey1}. 
This approach significantly improves computational efficiency while achieving higher evaluation accuracy compared to traditional Monte Carlo simulation-based methods.
Based on DiSE, we introduce a training-free flexible-length sequence generation method that, unlike conventional fixed-length generation, enables controllable and adaptive output lengths guided by the model's self-assessment. 
Serving as a real-time self-evaluation mechanism, DiSE guides the process of searching, assessing and stopping to determine the optimal generation length.
Extensive experiments on likelihood evaluation, uncertainty quantification, and flexible-length generation show the effectiveness of the proposed DiSE.

\section{DiSE}

\subsection{Preliminary: dLLM Monte Carlo Probability Estimation}

DLLMs do not employ the causal masking used in auto-regressive LLMs and therefore the probability of generating a sequence cannot be factorized as a simple product of conditional probabilities. To approximate the log-probability of generating a target sequence $X^0 = (x^0_1, x^0_2, \dots, x^0_N)$, the traditional approach~\citep{llada} adopts the following term:

\begin{small}
\begin{equation}
\mathbb{E}_{l, X^l}\left[\frac{N}{l} \sum_{i=1}^{N} \mathbf{1}\left[x^l_i= \langle \mathrm{mask\ token} \rangle \right] \log p_\theta\left(x^0_i \mid X^l\right)\right],
\end{equation}
\end{small}
where $l$ is uniformly sampled from $\{1, 2, \dots, N\}$, and $X^l=(x^l_1, x^l_2, \dots, x^l_N)$ is obtained by uniformly sampling $l$ tokens from $X^0$, replacing the tokens at these positions with mask tokens, while keeping all other tokens identical to those in $X^0$.
Since the exact computation of this expectation is intractable, Monte Carlo simulation~\citep{harrison2010introduction} is employed, where a finite number of samples are generated and the expectation is approximated by their empirical average. This approximation enables tractable estimation of sequence probabilities for dLLMs. 
The probability estimation for conditional generation and auto-regressive LLM probability estimation is detailed in Appendix~\ref{sec:more-probability-estimation-formulas}.

\subsection{Definition}

\begin{figure*}[t]
    \centering
    \includegraphics[width=\linewidth]{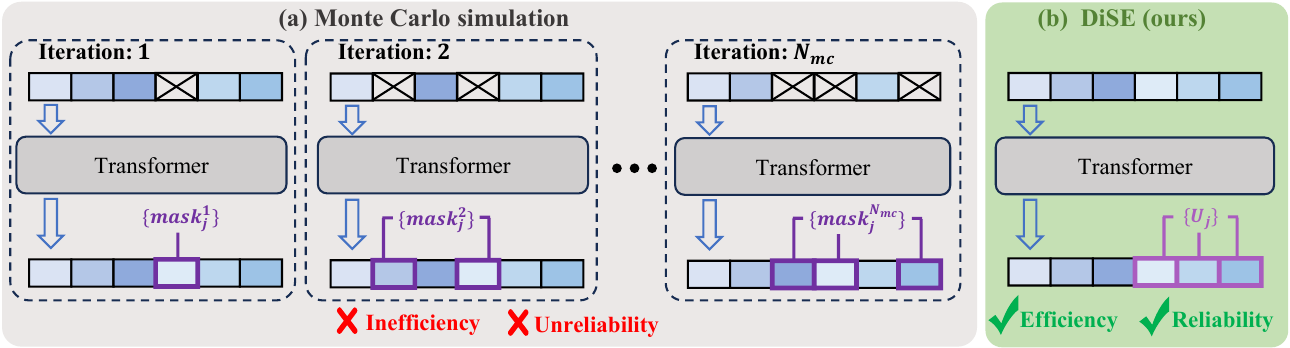}
    \vspace{-8mm}
    \caption{A simplified illustration of self-evaluation confidence quantification methods for clarity. 
    (a) Monte Carlo simulation approach for dLLMs. A total of $N_{mc}$ simulations are performed. In the $i$-th simulation, a set of masked positions $\{mask^i_j\}$ is sampled. The tokens at these positions are replaced with mask tokens, and the model predicts the probability of correctly generating these tokens. The final estimation is obtained by aggregating the results across all $N_{mc}$ simulations.
    (b) The proposed DiSE for dLLMs. The set of selected positions $\{U_j\}$ is predefined. The model receives the entire sequence and estimates the regeneration probability of the tokens at $\{U_j\}$. 
    }
    
    \label{fig:overview}
\end{figure*}

In traditional likelihood estimation approaches, whether using auto-regressive LLMs or dLLMs with Monte Carlo simulation, the common paradigm is to condition on the tokens at known positions and predict the tokens at unknown positions based on their probability distributions.
However, under the dLLM framework, it is also possible to predict the tokens at positions that are already known. 
In this work, we propose DiSE, a self-evaluation confidence quantification method for dLLMs that employs token regeneration probability as a novel indicator of model confidence and investigate different token sets to calculate token regeneration probability. 


Let the text sequence be $X = (x_1, x_2, \dots, x_N)$. The dLLM takes $X$ as input and concurrently predicts the tokens at all positions that already exist. $p_\theta(x_i \mid X)$ represents the probability of the model regenerating token $x_i$ at position $i$ given the entire sequence $X$. Accordingly, the probability of the model regenerating $X$ given $X$ is formulated as $\prod_{i=1}^{N} p_\theta(x_i \mid X).$
Consider a binary mask $M \in \{0,1\}^N$, where $M_i = 1$ indicates that the token at position $i$ is included in the probability calculation for regeneration, and $M_i = 0$ means it is ignored.
Let $U=\{ i \mid M_i = 1 \}$ be the index set of the selected positions. The probability of regenerating the tokens in the selected region is formulated as $\prod_{i \in U} p_\theta(x_i \mid X)$.
After taking the logarithm and averaging over the number of selected tokens, the DiSE score is defined as follows:
\begin{equation}
\mathrm{DiSE}(X) = \frac{1}{|U|} \sum_{i \in U} \log p_\theta(x_i \mid X),
\end{equation}
where different selection modes are employed to determine the binary mask $M \in \{0,1\}^N$, thereby controlling the index set of selected positions $U$.
This measure captures the model's confidence in regenerating its own tokens and allows flexible evaluation over either local regions or the entire sequence.
For conditional generation with prompt $P$ and generated response $R$, the DiSE score is calculated by treating the concatenated sequence $[P; R]$ as $X$.
Figure~\ref{fig:overview} presents a simplified visualization of the Monte Carlo simulation approach for dLLMs and the proposed DiSE.

\subsection{Analysis}

\begin{figure}[ht]
    \centering
    \includegraphics[width=\linewidth]{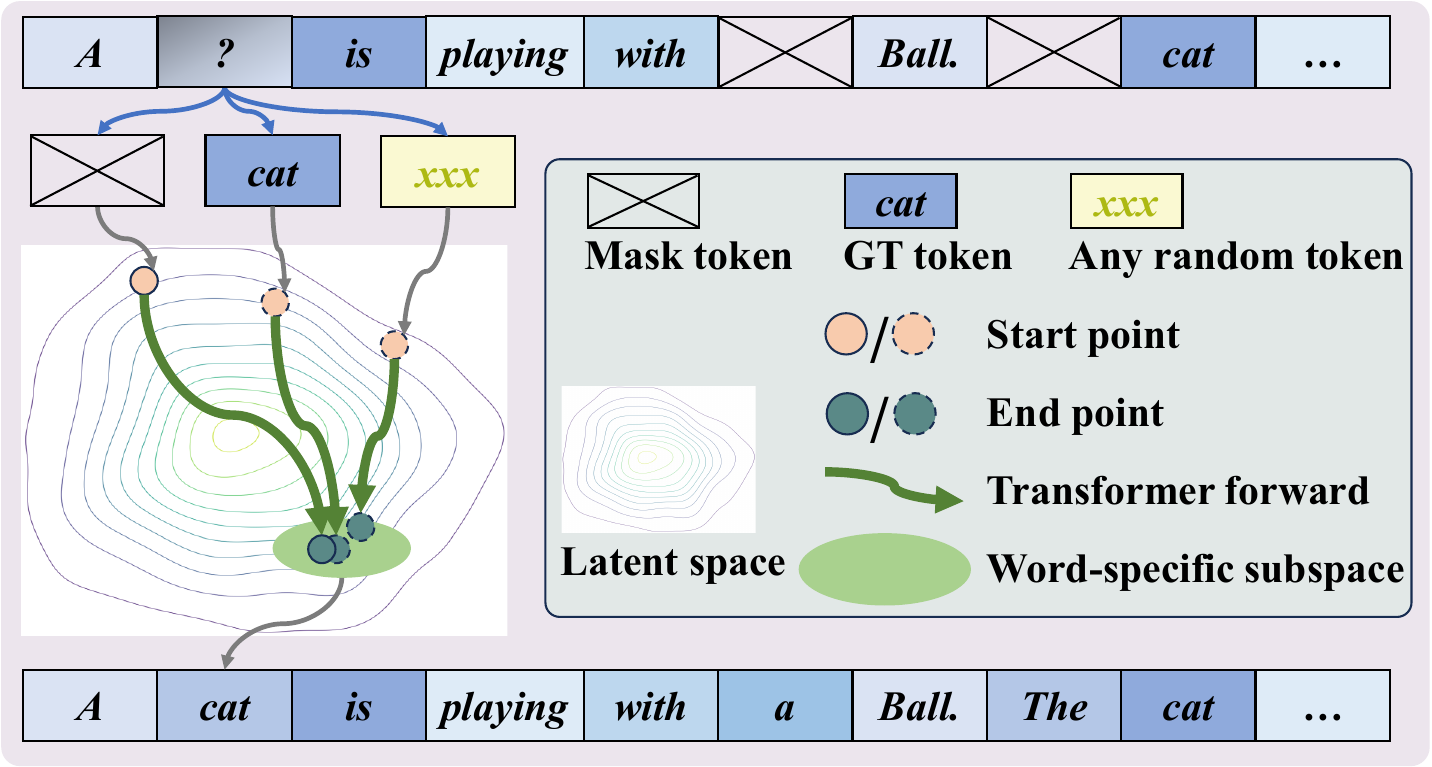}
    \vspace{-8mm}
    \caption{Generalization ability of dLLMs: Different tokens map from distinct start points to similar end points in the latent space.}
    \label{fig:method-analysis}
\end{figure}

During training, dLLMs receive no supervision for regenerating tokens already known in the input, meaning our method relies on a behavior the model is never explicitly taught. The interpretability of our approach therefore stems from the inherent generalization capability of dLLMs.

\noindent \textbf{Generalization Ability of dLLMs under Random Perturbations.}
As illustrated in Figure~\ref{fig:method-analysis}, consider a noisy sequence $\overline{X}$, where $\overline{x}_i$ is a mask token. $\overline{x}_i$ is first mapped to a start point in the latent space. After interacting with the surrounding context through transformer layers, it eventually moves toward an end point that falls within the word-specific subspace of the ground-truth (GT) token.
Through dLLM training, the network learns the ability to reach the correct word-specific subspace interacting with the surrounding context. When we replace $\overline{x}_i$ with a random token, although the starting point in the latent space changes, the model still tends to move toward the correct target subspace.

\begin{figure}[ht]
    \centering
    \includegraphics[width=\linewidth]{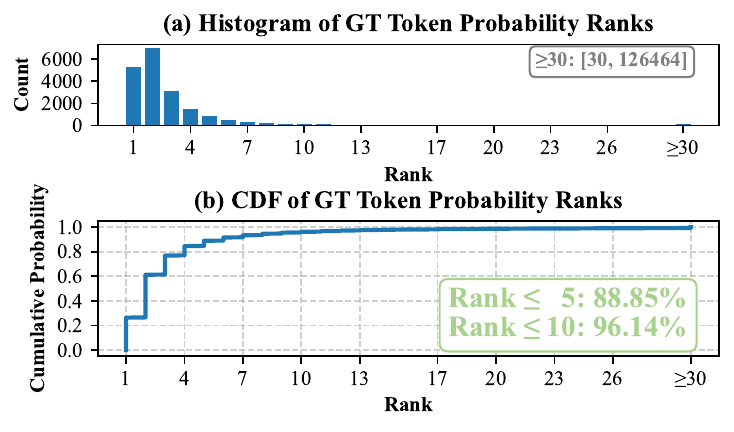}
    \vspace{-10mm}
    \caption{Histogram and Cumulative distribution function (CDF) of GT token probability ranks.}
    \label{fig:statistic-GT-rank-for-random-token-gather}
\end{figure}

Consider $\overline{X}^{R}_i$, which is derived from a complete sequence by replacing the token at position $i$ with a random token. Using $\overline{X}^{R}_i$ as input, we evaluate the rank of the GT token within the predicted probability distribution at position $i$. This procedure is repeated across multiple sentences, positions, and random token samples, resulting in a total of 20,000 trials. The distribution of GT token ranks is summarized in Figure~\ref{fig:statistic-GT-rank-for-random-token-gather}. Notably, the vast majority of GT tokens occupy top ranks, with an $88.85\%$ probability of ranking $\leq 5$ and a $96.14\%$ probability of ranking $\leq 10$, confirming the generalization capability of dLLMs from start points that are not encountered during training.

\noindent \textbf{GT Tokens Exhibit Better Generalization than Random Tokens.}
As illustrated in Figure~\ref{fig:method-analysis}, compared with a random token, the GT token naturally possesses semantic consistency with the surrounding context, enabling the model to reach the correct subspace more reliably and further enhancing the effectiveness of token regeneration.

\begin{figure}[ht]
    \centering
    \includegraphics[width=\linewidth]{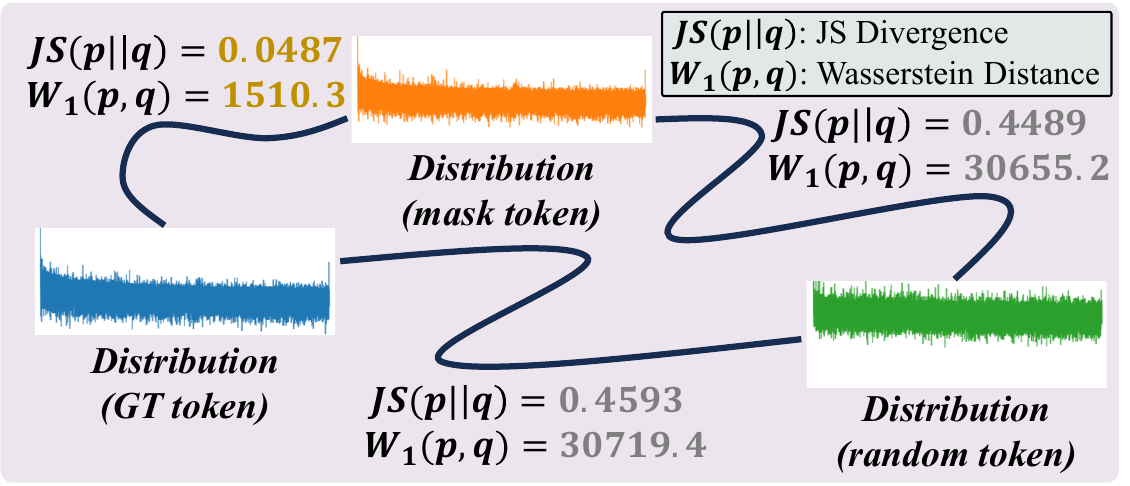}
    \vspace{-8mm}
    \caption{Mean pairwise distribution distances for GT, mask, and random tokens using JS Divergence and Wasserstein Distance.}
    \label{fig:distribution-distance}
\end{figure}


Similarly to $\overline{X}^{R}_i$, we define $\overline{X}^{G}_i$ and $\overline{X}^{M}_i$ as sequences obtained from a complete sentence by replacing the token at position $i$ with the GT token and the mask token, respectively. 
We feed $\overline{X}^{G}_i$, $\overline{X}^{M}_i$ and $\overline{X}^{R}_i$ into the model to obtain their respective predicted distributions at position $i$.
We sample a total of 2,509 instances and compute the pairwise distribution distances for each instance using JS Divergence and Wasserstein Distance, subsequently calculating the mean values. As shown in Figure~\ref{fig:distribution-distance}, the mean distribution distance between GT and mask tokens is significantly smaller than that between random and mask tokens, confirming the effectiveness of the token regeneration.
More analyses are detailed in Appendix~\ref{sec:more-analyses}.

\subsection{Observation}

\begin{figure}[ht]
    \centering
    \includegraphics[width=\linewidth]{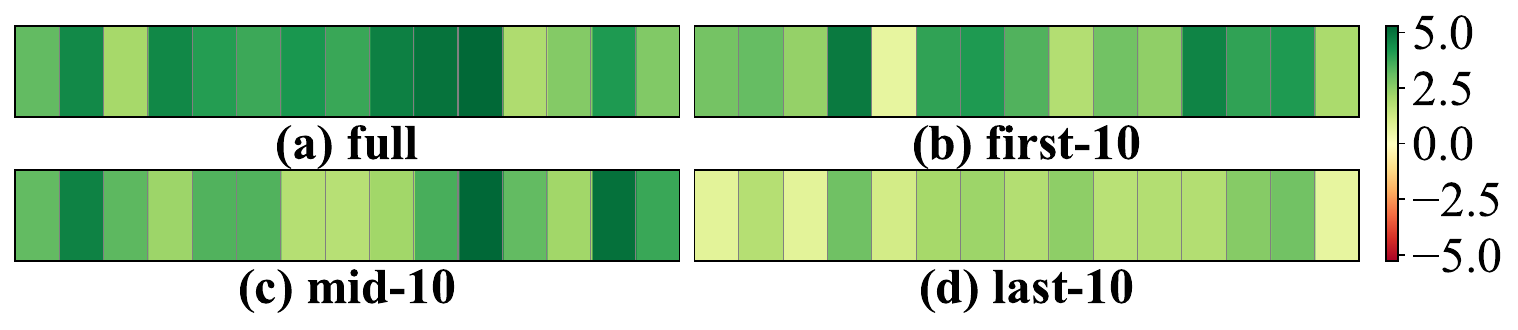}
    \vspace{-8mm}
    \caption{Differences between the DiSE scores of natural sentences and randomized sentences using the LLaDA-Instruct-8B model under four selection modes: `full' (entire sentence), `first-10' (first 10 tokens), `mid-10' (10 tokens from the middle) and `last-10' (last 10 tokens). Each subfigure contains 15 blocks, representing 15 sampled sentences. All blocks are shown in green (difference $>0$), indicating that natural sentences consistently achieve higher DiSE scores than randomized sentences.}
    \label{fig:observation-normal-random}
\end{figure}

\noindent \textbf{Observation I: Semantic Coherence Positively Correlates with DiSE Scores.}
%
%
We sample 15 well-formed sentences and generate fully randomized versions by replacing all original tokens with random tokens. The DiSE scores are computed for both the natural and randomized sentences using a binary mask $M$ with all positions set to one, corresponding to the selection mode `full'. As shown in Figure~\ref{fig:observation-normal-random} (a), natural sentences achieve substantially higher DiSE scores than their randomized counterparts. Additionally, we perform three local token randomization experiments, replacing 10 tokens in the front, middle or back regions of each sentence, and the DiSE scores are measured for these perturbed positions. In these experiments, the selection modes are denoted as `first-10'/`mid-10'/`last-10', indicating that $M=1$ is applied only to the respective region. Figures~\ref{fig:observation-normal-random} (b), (c) and (d) show that natural sentences consistently obtain higher DiSE scores than randomized sentences in all regions. These findings indicate that DiSE effectively captures semantic coherence of both global and local regions, allowing fine-grained self-evaluation across different parts of a sentence.

\begin{figure}[ht]
    \centering
    \includegraphics[width=\linewidth]{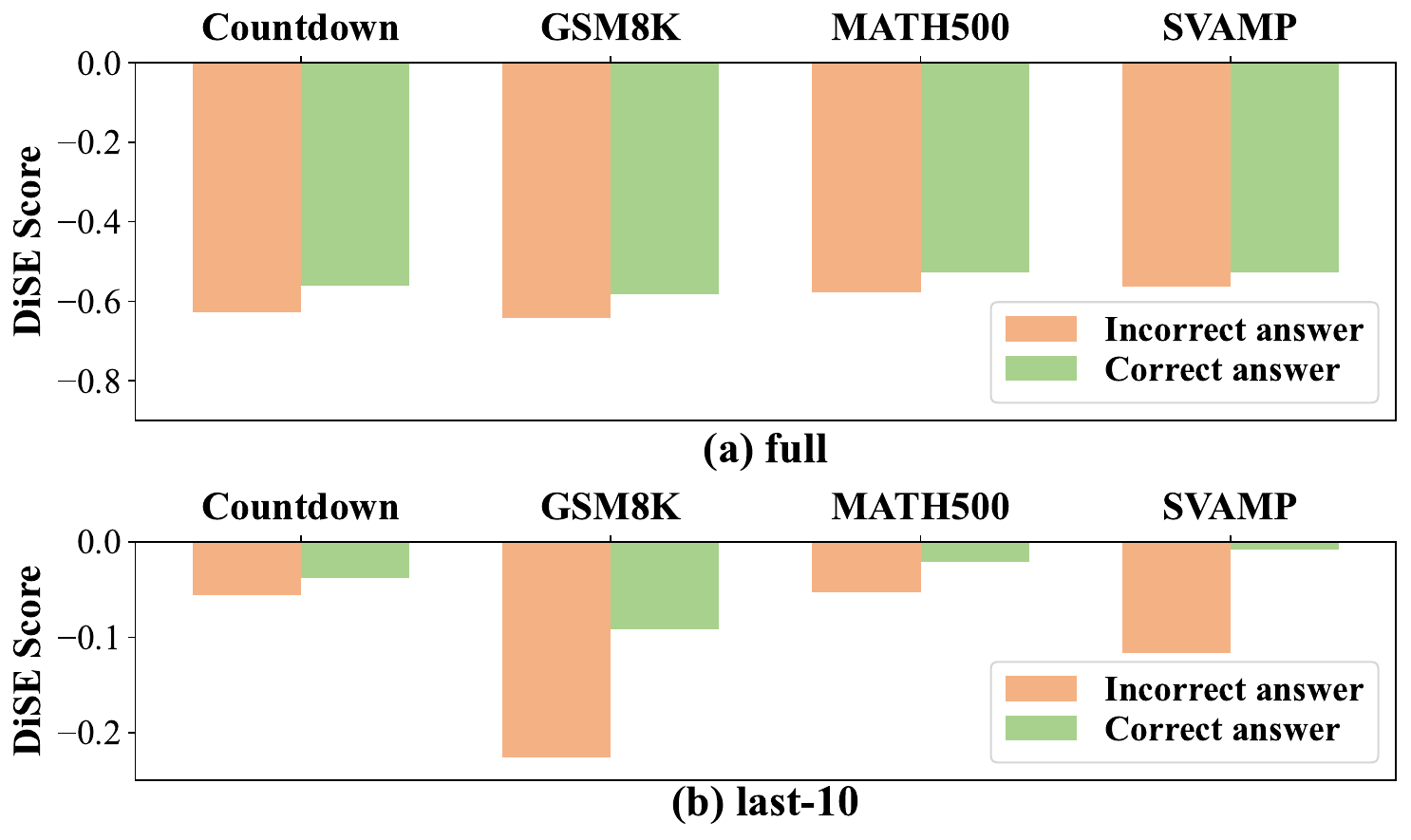}
    \vspace{-8mm}
    \caption{Comparison between the DiSE scores of correct and incorrect answers across four datasets using the LLaDA-Instruct-8B model under two selection modes `full' and `last-10'. 
    }
    \label{fig:observation-correct-incorrect-llada-instruct}
    \vspace{-4mm}
\end{figure}

\noindent \textbf{Observation II: Answer Accuracy Positively Correlates with DiSE Scores.}
%
We conduct a series of experiments on four commonly used reasoning datasets: Countdown~\citep{dataset-countdown}, GSM8K~\citep{dataset-gsm8k}, MATH500~\citep{dataset-math500} and SVAMP~\citep{dataset-svamp}. The model outputs are categorized into two groups according to whether the generated answers match the ground-truth solutions. We compute the DiSE scores separately for the correct and incorrect groups and report their averages under two selection modes `full' and `last-10'. The results, summarized in Figure~\ref{fig:observation-correct-incorrect-llada-instruct}, consistently reveal that correct outputs tend to exhibit higher DiSE scores than incorrect ones across different datasets. 
Importantly, under the selection mode `last-10', which focuses on the final ten tokens closely associated with the answer positions, the disparity between correct and incorrect outputs is substantially amplified. 
This finding highlights the strong correlation between DiSE scores and answer accuracy, supporting the reliability of the proposed DiSE.

\section{Applying DiSE in Real-World Scenarios}


\subsection{Conditional Likelihood Estimation for dLLMs.}
Conditional likelihood estimation serves as an important metric for evaluating the generative ability of language models.
During the evaluation, we estimate the probability or log-probability of generating a candidate response $R$ conditioned on a given prompt $P$. 
For each prompt $P$, there may be multiple candidate responses, and we select the one with the highest probability as the final answer and compute the accuracy accordingly.
In this work, DiSE is employed as an approximate estimator of the conditional likelihood evaluation via the unconventional regeneration probability, rather than the standard generation probability.

\subsection{Uncertainty Quantification for dLLMs.}
Quantifying the uncertainty of model outputs is crucial for assessing their reliability. In the context of dLLMs, we use the DiSE score as a self-evaluation signal to measure the confidence of a generated sequence. 
Sequences with higher DiSE scores are considered more reliable, while lower scores indicate higher uncertainty. 
The negative of the DiSE score is used to quantify the uncertainty of the model output, with a higher value reflecting a higher estimated uncertainty.


\subsection{Flexible-length dLLM Generation with DiSE}

\begin{figure*}[t]
    \centering
    \includegraphics[width=\linewidth]{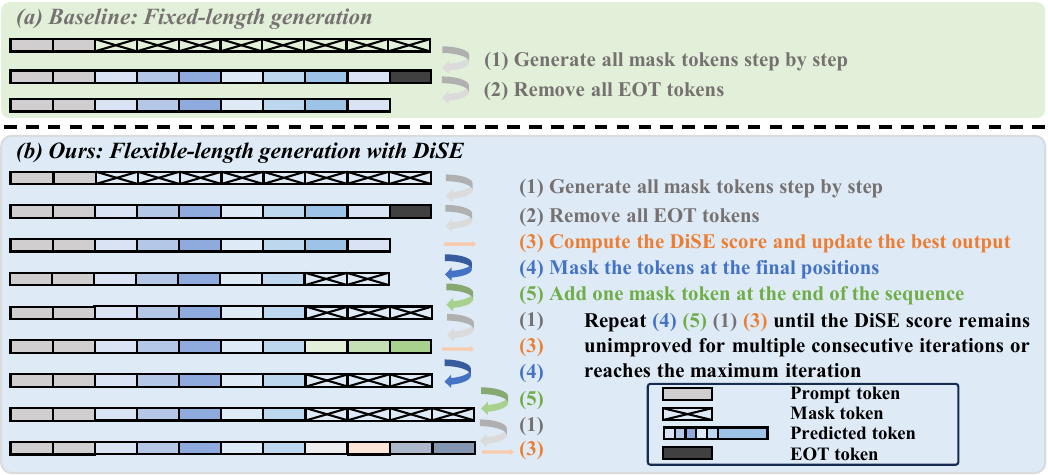}
    \vspace{-8mm}
    \caption{Illustration of the flexible-length dLLM generation framework with DiSE versus fixed-length generation. 
    (a) Fixed-length generation baseline. 
    (b) Flexible-length generation with DiSE. 
    }
    \label{fig:flexible-length-gen}
\end{figure*}

In general, dLLMs require the generation length $L$ to be fixed and specified in advance. Different choices of $L$ lead to different outcomes, and longer generations incur higher computational costs. 
In our work, we aim to relax the restriction of a fixed generation length and instead allow the output length to be adjusted flexibly within a controllable range. 
This is enabled by DiSE, which provides an intrinsic signal to evaluate the quality of generations without ground-truth supervision. Leveraging this property, we propose a training-free flexible-length dLLM generation method with DiSE.

Our method proceeds as follows. 
Given a prompt $P$ and a base length $L$, we first generate an initial response $R$ of length $L$. Let $\overline{R}$ denote the sequence obtained by removing all EOT tokens from $R$. We construct the complete token sequence as $X^{(1)} = [P; \overline{R}]$ and compute its DiSE score, which serves as the guiding criterion for controlling the generation length.
Keeping the tokens in the early positions unchanged, we apply a masking operation to the last $D$ tokens, and add one additional mask token at the end of the sequence. We use the model to regenerate the sequence, after which the DiSE score of the newly generated sequence is computed. At each iteration, $D$ is incremented by one. This process is repeated iteratively, with DiSE determining whether the extended generation is beneficial. If the DiSE score improves, we retain the extension; otherwise, if the DiSE score remains unimproved for $K$ consecutive iterations, we stop.
To avoid unbounded computation, we set a maximum of $M_{max}$ iterations. 
The overall procedure is illustrated in Figure~\ref{fig:flexible-length-gen}.
This flexible-length generation process uses the DiSE score as a self-evaluation signal, enabling dLLMs to adaptively decide their output length in a principled manner.
The detailed algorithm is provided in Appendix~\ref{sec:flexible-generation-algorithm}.
\section{Experiments}

\subsection{Experimental Setup}
\label{sec:experimental-setup}


Experiments are conducted using two dLLMs, LLaDA-Instruct-8B~\citep{llada} and LLaDA-1.5-8B~\citep{llada1.5}, on a diverse set of datasets, including ARC-Challenge~\citep{dataset-arc-challenge}, GPQA~\citep{dataset-gpqa}, Countdown~\citep{dataset-countdown}, GSM8K~\citep{dataset-gsm8k}, MATH500~\citep{dataset-math500} and SVAMP~\citep{dataset-svamp}.
The conventional Monte Carlo simulation approach for dLLMs is used as the baseline, with the number of samples $N_{mc}$ evaluated under two settings: $N_{mc}=1$ and $N_{mc}=32$. 
Additionally, we include the auto-regressive LLM LLaMA3-Instruct-8B~\citep{llama3} for comparison in the experiments.
More details are presented in Appendix~\ref{sec:more-implementation-details}.

\subsection{Conditional Likelihood Estimation}


We evaluate our approach on conditional likelihood estimation, with results summarized in Table~\ref{tab:conditional-likelihood-estimation}. Compared to the conventional Monte Carlo baseline, our method demonstrates substantial and consistent gains on ARC-Challenge and GPQA, demonstrating its reliability as a likelihood estimator. It also achieves comparable or superior accuracy to auto-regressive LLM probability estimates. In addition, we report the average number of model forward passes per computation. Notably, compared with Monte Carlo sampling at $N_{mc}=32$, our method achieves nearly a $32\times$ efficiency improvement while providing higher accuracy. 
Using LLaDA-Instruct-8B, our approach outperforms the $N_{mc}=1$ Monte Carlo baseline (with similar computational cost) by $23.6\%$ on ARC-Challenge and $8.9\%$ on GPQA. Even against the more expensive $N_{mc}=32$ baseline, it delivers both a $32\times$ speedup and improved accuracy, with gains of $6.4\%$ and $1.5\%$ respectively. Additional results for Dream-Instruct-7B~\citep{dream2025} are presented in Appendix~\ref{sec:additional-results-using-dream7B}.

\begin{table}[t] 
\centering
\caption{Conditional likelihood estimation results on ARC-Challenge and GPQA. The table compares the proposed DiSE against the Monte Carlo simulation baseline with varying $N_{mc}$, and also includes a comparison with the probability estimates from auto-regressive LLMs. The last column reports the average number of model forward passes per computation.}
\label{tab:conditional-likelihood-estimation} 
\vspace{-4mm}
\resizebox{\linewidth}{!}{
    \begin{tabular}{ll|cc|c}
        \toprule
        & \textbf{Method} & \textbf{ARC-Challenge} & \textbf{GPQA} & \textbf{\# NFE}  \\
        \midrule
        \multirow{4}{*}{\textbf{LLaDA-Instruct-8B}} 
        & MC, $N_{mc}=1$ & 0.306 & 0.212 & 1 \\
        \cmidrule(lr){2-5}
        & MC, $N_{mc}=32$ & 0.478 & 0.286 & 32 \\
        \cmidrule(lr){2-5}
        & \textbf{DiSE (ours)} & \textbf{0.542} & \textbf{0.301} & 1 \\
        \midrule
        \midrule
        \multirow{4}{*}{\textbf{LLaDA-1.5-8B}} 
        & MC, $N_{mc}=1$ & 0.311 & 0.203 & 1 \\
        \cmidrule(lr){2-5}
        & MC, $N_{mc}=32$ & 0.488 & 0.275 & 32 \\
        \cmidrule(lr){2-5}
        & \textbf{DiSE (ours)} & \textbf{0.567} & \textbf{0.299} & 1 \\
        \midrule
        \midrule
        \textcolor{gray}{\textbf{LLaMA-3-8B}} & \textcolor{gray}{probability} & \textcolor{gray}{0.530} & \textcolor{gray}{0.304} & \textcolor{gray}{1}  \\
        \bottomrule
    \end{tabular}
}
\vspace{-2mm}
\end{table}

\begin{table*}[t] 
\centering
\caption{ROC-AUC scores for uncertainty quantification on the Countdown, GSM8K, MATH500 and SVAMP datasets with varing generation lengths. The table compares Monte Carlo simulation baseline with varying $N_{mc}$, the proposed DiSE, and the perplexity calculation using the auto-regressive model LLaMA3-Instruct-8B. The last column reports the average ROC-AUC scores across the preceding 12 settings.}
\label{tab:ROC-AUC} 
\vspace{-4mm}
\resizebox{\textwidth}{!}{
    \begin{tabular}{ll|cccccccccccc|>{\columncolor{yellow!20}}c}
        \toprule
        &
        & \multicolumn{3}{c}{\textbf{Countdown}}
        & \multicolumn{3}{c}{\textbf{GSM8K}} 
        & \multicolumn{3}{c}{\textbf{MATH500}} 
        & \multicolumn{3}{c|}{\textbf{SVAMP}} 
        & \multirow{2}{*}{\textbf{Avg. ROC-AUC$\uparrow$}} \\
        \cmidrule(lr){3-5} \cmidrule(lr){6-8} \cmidrule(lr){9-11} \cmidrule(lr){12-14}
         & \textbf{Method / Gen Len} & 128 & 256 & 512 & 128 & 256 & 512 & 128 & 256 & 512 & 128 & 256 & 512 & \\
        \midrule
        \multirow{5}{*}{\textbf{LLaDA-Instruct-8B}} 
        & MC, $N_{mc}=1$ & 0.524 & 0.520 & 0.528 & 0.539 & 0.513 & 0.540 & 0.497 & 0.541 & 0.532 & 0.563 & 0.575 & 0.509 & 0.532 \\
        \cmidrule(lr){2-15}
        & MC, $N_{mc}=32$ & \textbf{0.595} & \textbf{0.534} & 0.558 & 0.590 & 0.552 & 0.595 & 0.528 & 0.578 & 0.531 & 0.616 & 0.551 & 0.647 & 0.573 \\
        \cmidrule(lr){2-15}

        
        & \textbf{DiSE (ours)} & 0.578 & 0.521 & \textbf{0.622} & \textbf{0.633} & \textbf{0.644} & \textbf{0.658} & \textbf{0.611} & \textbf{0.634} & \textbf{0.604} & \textbf{0.688} & \textbf{0.692} & \textbf{0.755} & \textbf{0.637} \\

        \cmidrule(lr){2-15}
        \morecmidrules
        \cmidrule(lr){2-15}
        & \textcolor{gray}{LLaMA perplexity} & \textcolor{gray}{0.574} & \textcolor{gray}{0.419} & \textcolor{gray}{0.392} & \textcolor{gray}{0.675} & \textcolor{gray}{0.605} & \textcolor{gray}{0.577} & \textcolor{gray}{0.575} & \textcolor{gray}{0.637} & \textcolor{gray}{0.551} & \textcolor{gray}{0.686} & \textcolor{gray}{0.650} & \textcolor{gray}{0.590} & \textcolor{gray}{0.578} \\
        \midrule
        \midrule
        \multirow{5}{*}{\textbf{LLaDA-1.5-8B}} 
        & MC, $N_{mc}=1$ & 0.525 & \textbf{0.588} & 0.528 & 0.516 & 0.559 & 0.525 & 0.558 & 0.514 & 0.525 & 0.562 & 0.466 & 0.554 & 0.535\\
        \cmidrule(lr){2-15}
        & MC, $N_{mc}=32$ & 0.608 & 0.557 & 0.520 & 0.559 & 0.578 & 0.608 & 0.580 & 0.546 & \textbf{0.551} & 0.585 & 0.513 & 0.597 & 0.567\\
        \cmidrule(lr){2-15}
        
        & \textbf{DiSE (ours)} & \textbf{0.610} & 0.471 & \textbf{0.586} & \textbf{0.610} & \textbf{0.616} & \textbf{0.613} & \textbf{0.606} & \textbf{0.553} & 0.533 & \textbf{0.599} & \textbf{0.629} & \textbf{0.677} & \textbf{0.592} \\

        \cmidrule(lr){2-15}
        \morecmidrules
        \cmidrule(lr){2-15}
        & \textcolor{gray}{LLaMA perplexity} & \textcolor{gray}{0.596} & \textcolor{gray}{0.459} & \textcolor{gray}{0.362} & \textcolor{gray}{0.635} & \textcolor{gray}{0.631} & \textcolor{gray}{0.546} & \textcolor{gray}{0.652} & \textcolor{gray}{0.588} & \textcolor{gray}{0.550} & \textcolor{gray}{0.639} & \textcolor{gray}{0.587} & \textcolor{gray}{0.620} & \textcolor{gray}{0.572} \\
        \bottomrule
    \end{tabular}
}
\end{table*}

\subsection{Uncertainty Quantification}
\label{sec:experiments-uncertainty-quantification}


For uncertainty quantification experiments, we evaluate the ability to distinguish correctness among multiple generated answers for each question using ROC-AUC scores~\citep{kuhn2023semantic}, where the ROC-AUC score measures the probability that a randomly chosen correct answer receives lower uncertainty than a randomly chosen incorrect one. We generate $5$ answers per question.
Table~\ref{tab:ROC-AUC} reports results on Countdown, GSM8K, MATH500, and SVAMP with varying generation lengths. Compared with the conventional Monte Carlo approach, our method achieves substantial improvements. 
Using LLaDA-Instruct-8B, it improves average ROC-AUC by $10.5\%$ over Monte Carlo with $N_{mc}=1$ at comparable cost. Even compared to Monte Carlo with $N_{mc}=32$, which incurs a nearly $32\times$ higher cost, our approach remains superior by $6.4\%$. Compared with the perplexity-based uncertainty from an auto-regressive LLM, our method yields a $5.9\%$ gain on the same generations. Additional best-of-N sampling results are presented in Appendix~\ref{sec:best-of-n-results}.

\begin{figure}[t]
    \centering
    \includegraphics[width=\linewidth]{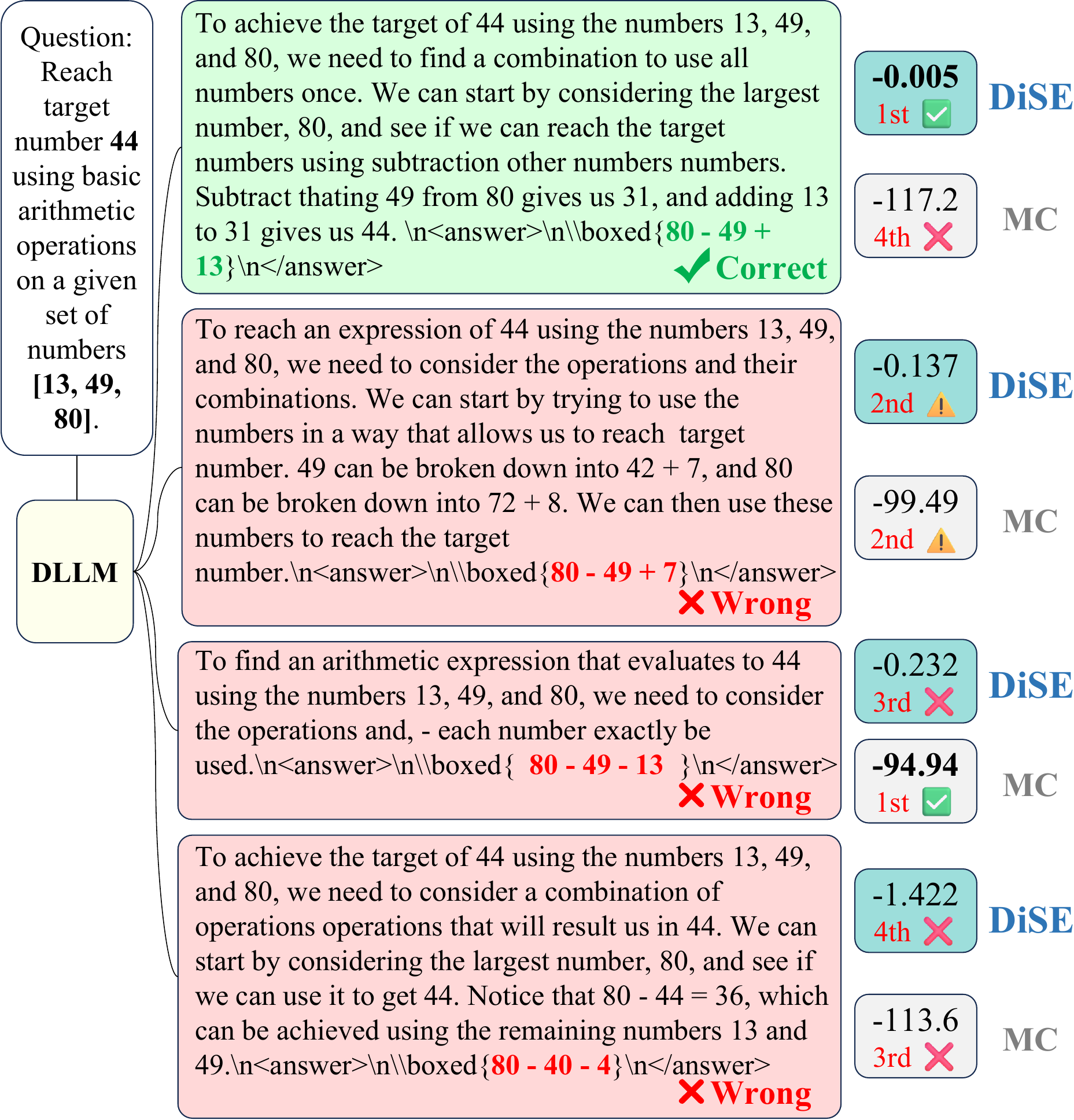}
    \vspace{-8mm}
    \caption{Qualitative example of uncertainty quantification with four generated answers using LLaDA-Instruct-8B.
    }
    \label{fig:qualitative-results}
\end{figure}

We present a qualitative example in Figure~\ref{fig:qualitative-results} comparing DiSE with Monte Carlo simulation in capturing answer correctness.
Using four candidate answers generated by LLaDA-Instruct-8B from the same input, DiSE consistently assigns lower scores to incorrect answers, corresponding to higher uncertainty, while Monte Carlo with $N_{mc}=32$ fails to do so.
This example demonstrates that DiSE offers more reliable, fine-grained sequence-level uncertainty estimates. 
Additional qualitative examples are presented in Appendix~\ref{sec:additional-example-uncertainty-quantification}.

\begin{figure}[t]
    \centering
    \includegraphics[width=\linewidth]{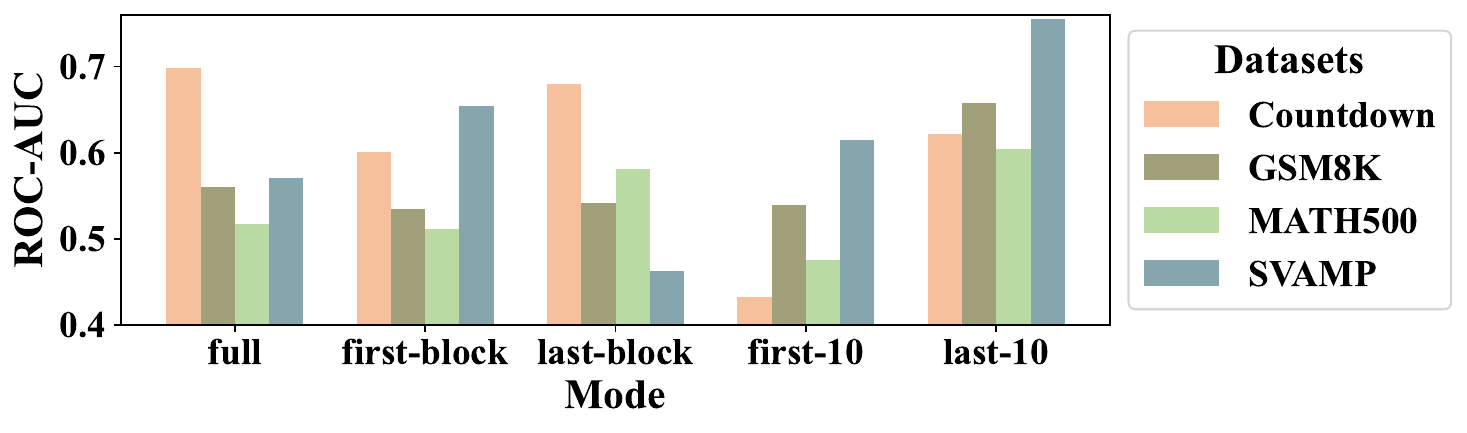}
    \vspace{-8mm}
    \caption{Ablation study of different DiSE selection modes for uncertainty quantification using LLaDA-Instruct-8B with a generation length of 512. 
    }
    \label{fig:DiSE-ablation}
\end{figure}

We investigate the effect of different DiSE selection modes on uncertainty quantification, where each mode specifies the subset of tokens used for computing regeneration probability: `full' (all tokens), `first-block' (tokens in the first generation block), `last-block' (tokens in the last generation block, including EOT tokens), `first-10' (first 10 generated tokens) and `last-10' (last 10 non-EOT tokens).
Figure~\ref{fig:DiSE-ablation} shows that using the last 10 non-EOT tokens tends to yield higher ROC-AUC scores on multiple datasets, as these tokens typically correspond to the answer region. 
Earlier tokens offer limited information on correctness, and including EOT tokens in the last block degrades performance, which aligns with our intuition.
Additional results under different selection modes are presented in Appendix~\ref{sec:selection-mode-ablation-more}.

\subsection{Flexible-length dLLM Generation}


\begin{table*}[t] 
\centering
\caption{Results of flexible-length dLLM generation with DiSE on the Countdown, GSM8K, MATH500, and SVAMP datasets with varying base lengths. 
The table shows two fixed-length baselines: \textbf{Baseline}, generating sequences of base length $L$, and \textbf{Baseline (Max Len)}, generating sequences of length $L + M_{max}$.
These are compared with the proposed flexible-length generation with DiSE (DiSE-flexible). The final column reports the average accuracy across the preceding 12 configurations.}
\label{tab:fleixble-length-generation} 
\vspace{-4mm}
\resizebox{\textwidth}{!}{
    \begin{tabular}{ll|cccccccccccc|>{\columncolor{yellow!20}}c}
        \toprule
        &
        & \multicolumn{3}{c}{\textbf{Countdown}}
        & \multicolumn{3}{c}{\textbf{GSM8K}} 
        & \multicolumn{3}{c}{\textbf{MATH500}} 
        & \multicolumn{3}{c|}{\textbf{SVAMP}} 
        & \multirow{2}{*}{\textbf{Avg. Accuracy}} \\
        \cmidrule(lr){3-5} \cmidrule(lr){6-8} \cmidrule(lr){9-11} \cmidrule(lr){12-14}
         & \textbf{Method / Base Len} & 128 & 256 & 512 & 128 & 256 & 512 & 128 & 256 & 512 & 128 & 256 & 512 & \\
        \midrule
        \multirow{6}{*}{\textbf{LLaDA-Instruct-8B}} 
        & Baseline   & 26.17 & 15.23 & 12.50 & 68.01 & 76.65 & 79.23 & \textbf{26.20} & 32.80 & \textbf{36.80} & 84.67 & 85.00 & 83.67 & 52.24 \\
        \cmidrule(lr){2-15}
        & Baseline (Max Len) & 25.00 & 16.41 & \textbf{15.62} & 69.29 & 76.80 & 78.85 & 25.60 & 31.60 & 36.40 & 85.33 & 84.67 & 83.00 & 52.38 \\
        \cmidrule(lr){2-15}

        
        & \textbf{DiSE-flexible (ours)} & \textbf{27.73} & \textbf{18.36} & \textbf{15.62} & \textbf{70.96} & \textbf{79.68} & \textbf{79.30} & 26.00 & \textbf{33.60} & 36.60 & \textbf{87.33} & \textbf{86.00} & \textbf{84.33} & \textbf{53.79} \\

        \midrule
        \midrule
        \multirow{6}{*}{\textbf{LLaDA-1.5-8B}} 
        & Baseline   & 24.22 & 15.62 & 17.19 & 70.51 & 77.48 & 79.53 & 26.80 & 34.00 & 36.80 & \textbf{87.00} & 84.67 & 86.67 & 53.37\\
        \cmidrule(lr){2-15}
        & Baseline (Max Len) & 24.22 & 17.58 & 17.58 & 71.95 & 78.77 & 79.53 & 25.80 & 34.20 & 37.00 & 86.33 & 83.00 & 86.33 & 53.52 \\
        \cmidrule(lr){2-15}
        
        & \textbf{DiSE-flexible (ours)} & \textbf{26.17} & \textbf{19.53} & \textbf{22.27} & \textbf{72.33} & \textbf{79.53} & \textbf{80.06} & \textbf{27.20} & \textbf{35.60} & \textbf{37.40} & \textbf{87.00} & \textbf{85.00} & \textbf{87.00} & \textbf{54.92} \\
        \bottomrule
    \end{tabular}
}
\vspace{-1mm}
\end{table*}

Table~\ref{tab:fleixble-length-generation} presents the evaluation results of flexible-length dLLM generation on the Countdown, GSM8K, MATH500 and SVAMP datasets with multiple base lengths $L$. 
Two fixed-length baselines, generating sequences of length $L$ or $L + M_{max}$, are considered to reflect conventional fixed-length generation.
In contrast, our proposed method employs DiSE to guide flexible-length generation, enabling adaptive adjustment of the output sequence length. 
The results indicate that the flexible-length approach with DiSE yields average improvements over fixed-length baselines across multiple datasets and varying base lengths, providing strong evidence for the effectiveness of dynamically adapting sequence length with DiSE in dLLM generation. 
Comparison with another training-free flexible-length dLLM generation method is presented in Appendix~\ref{sec:comparison-with-daedal}.
The ablation results are presented in Appendix~\ref{sec:flexible-generation-ablation-more} and comparison with another training-free flexible-length dLLM generation method is presented in Appendix~\ref{sec:comparison-with-daedal}.


\section{Related Work}

\subsection{dLLMs}

Diffusion Large Language Models (dLLMs)~\citep{dLLM-survey} adapt the diffusion modeling paradigm~\citep{DDPM, IDDPM, DDIM}, which is originally successful in image and video generation~\citep{podell2023sdxl, zhong2025outdreamer}, to natural language. 
Early efforts, such as D3PM~\citep{d3pm}, DiffusionBERT~\citep{d3pm}, RDM~\citep{rdm}, MDLM~\citep{mdlm} and MD4~\citep{md4}, focused on exploring training objectives, noise scheduling strategies, and parameterization methods. 
Recent research includes LLaDA~\citep{llada}, the first large-scale dLLM, DIFFUSION-LLMs~\citep{diffusion-llms} with multi-stage training strategies, and DiffuGPT / DiffuLLaMA~\citep{DiffuGPT-DiffuLLaMA}, which adapt pre-trained auto-regressive models to the diffusion framework. DREAM~\citep{dream2025} further demonstrates strong performance in complex reasoning tasks. Subsequent developments, such as LLaDA 1.5~\citep{llada1.5} with variance-reduced preference optimization for preference alignment and TESS 2~\citep{tess-2} with auto-regressive initialization and adaptive noise scheduling, further improve generation quality.

\subsection{Self-Evaluation for LLMs}

Self-evaluation~\citep{ren2023self, geng2023survey} has emerged as a crucial mechanism in LLMs, providing models with the capability to assess the reliability of their own outputs and to produce internal measures of confidence and correctness.
Self-evaluation is most directly performed via likelihood estimation, using the model's probabilistic output to quantify plausibility. 
Beyond likelihoods, uncertainty quantification (UQ)~\citep{uq-survey1, uq-survey2, uncertainty-benchmarking} assesses model confidence and is essential for mitigating hallucinations in risk-sensitive applications. Token-level UQ estimates uncertainty from conditional probability distributions using entropy-based metrics, normalization schemes, or meaning-aware scores such as perplexity~\citep{uq-survey1}, CCP~\citep{fadeeva2024fact}, and MARS~\citep{bakman2024mars}. Self-verbalized UQ~\citep{stengel2024lacie, xu2024sayself, lin2022teaching} further elicits confidence through explicit probability statements or epistemic markers. Leveraging these signals, recent work employs self-evaluation for calibration to better align model confidence with empirical accuracy, improving the reliability of generated outputs~\citep{huang2024calibrating, xie2024calibrating}.

\section{Conclusion}


We introduce DiSE, a simple yet effective self-evaluation confidence quantification method for dLLMs. By employing token regeneration probability, DiSE achieves both high reliability and computational efficiency. 
Building upon DiSE, we propose a flexible-length generation framework, which enables adaptive sequence lengths through real-time self-evaluation.
Extensive analyses and validations confirm the feasibility of DiSE.
Comprehensive experiments across multiple datasets demonstrate the effectiveness of DiSE and the flexible-length generation framework with DiSE. 
DiSE closes the gap in dLLMs by introducing an efficient self-evaluation mechanism previously exclusive to auto-regressive LLMs. By leveraging DiSE, we overcome the fixed-length generation constraint in dLLMs and open the door to broader applications.

\section*{Limitations}

Semi-autoregressive models that integrate dLLMs with auto-regressive LLMs have recently emerged, but our method has not yet been evaluated on such architectures. Given the differences in training strategies and model design, their performance may differ, which we plan to investigate in future work. 
Moreover, although our current experiments achieve strong results using simple token selection strategies to compute DiSE, the optimal set of regeneration tokens for computation may vary across different tasks. Developing methods to systematically determine the best token subset for DiSE computation is a focus for future research.


\bibliography{main}

@article{dataset-gsm8k,
  title={Training verifiers to solve math word problems},
  author={Cobbe, Karl and Kosaraju, Vineet and Bavarian, Mohammad and Chen, Mark and Jun, Heewoo and Kaiser, Lukasz and Plappert, Matthias and Tworek, Jerry and Hilton, Jacob and Nakano, Reiichiro and others},
  journal={arXiv preprint arXiv:2110.14168},
  year={2021}
}

@inproceedings{dataset-math500,
  title={Let's verify step by step},
  author={Lightman, Hunter and Kosaraju, Vineet and Burda, Yuri and Edwards, Harrison and Baker, Bowen and Lee, Teddy and Leike, Jan and Schulman, John and Sutskever, Ilya and Cobbe, Karl},
  booktitle={The Twelfth International Conference on Learning Representations},
  year={2023}
}

@article{dataset-svamp,
  title={Are NLP models really able to solve simple math word problems?},
  author={Patel, Arkil and Bhattamishra, Satwik and Goyal, Navin},
  journal={arXiv preprint arXiv:2103.07191},
  year={2021}
}

@misc{dataset-countdown,
    author       = {Jiayi Pan and Junjie Zhang and Xingyao Wang and Lifan Yuan and Hao Peng and Alane Suhr},
    title        = {TinyZero},
    howpublished = {https://github.com/Jiayi-Pan/TinyZero},
    note         = {Accessed: 2025-01-24},
    year         = {2025}
}

@article{dataset-arc-challenge,
      author    = {Peter Clark  and Isaac Cowhey and Oren Etzioni and Tushar Khot and
                    Ashish Sabharwal and Carissa Schoenick and Oyvind Tafjord},
      title     = {Think you have Solved Question Answering? Try ARC, the AI2 Reasoning Challenge},
      journal   = {arXiv:1803.05457v1},
      year      = {2018},
}

@inproceedings{dataset-gpqa,
  title={Gpqa: A graduate-level google-proof q\&a benchmark},
  author={Rein, David and Hou, Betty Li and Stickland, Asa Cooper and Petty, Jackson and Pang, Richard Yuanzhe and Dirani, Julien and Michael, Julian and Bowman, Samuel R},
  booktitle={First Conference on Language Modeling},
  year={2024}
}

@article{llada,
  title={Large language diffusion models},
  author={Nie, Shen and Zhu, Fengqi and You, Zebin and Zhang, Xiaolu and Ou, Jingyang and Hu, Jun and Zhou, Jun and Lin, Yankai and Wen, Ji-Rong and Li, Chongxuan},
  journal={arXiv preprint arXiv:2502.09992},
  year={2025}
}

@article{llada1.5,
  title={LLaDA 1.5: Variance-Reduced Preference Optimization for Large Language Diffusion Models},
  author={Zhu, Fengqi and Wang, Rongzhen and Nie, Shen and Zhang, Xiaolu and Wu, Chunwei and Hu, Jun and Zhou, Jun and Chen, Jianfei and Lin, Yankai and Wen, Ji-Rong and others},
  journal={arXiv preprint arXiv:2505.19223},
  year={2025}
}

@misc{dream2025,
    title = {Dream 7B},
    url = {https://hkunlp.github.io/blog/2025/dream},
    author = {Ye, Jiacheng and Xie, Zhihui and Zheng, Lin and Gao, Jiahui and Wu, Zirui and Jiang, Xin and Li, Zhenguo and Kong, Lingpeng},
    year = {2025}
}

@article{llama3,
  title={The llama 3 herd of models},
  author={Dubey, Abhimanyu and Jauhri, Abhinav and Pandey, Abhinav and Kadian, Abhishek and Al-Dahle, Ahmad and Letman, Aiesha and Mathur, Akhil and Schelten, Alan and Yang, Amy and Fan, Angela and others},
  journal={arXiv e-prints},
  pages={arXiv--2407},
  year={2024}
}

@inproceedings{harrison2010introduction,
  title={Introduction to monte carlo simulation},
  author={Harrison, Robert L},
  booktitle={AIP conference proceedings},
  volume={1204},
  pages={17},
  year={2010}
}

@article{DDPM,
  title={Denoising diffusion probabilistic models},
  author={Ho, Jonathan and Jain, Ajay and Abbeel, Pieter},
  journal={Advances in Neural Information Processing Systems},
  volume={33},
  pages={6840--6851},
  year={2020}
}

@inproceedings{IDDPM,
  title={Improved denoising diffusion probabilistic models},
  author={Nichol, Alexander Quinn and Dhariwal, Prafulla},
  booktitle={International Conference on Machine Learning},
  pages={8162--8171},
  year={2021},
  organization={PMLR}
}

@article{DDIM,
  title={Denoising diffusion implicit models},
  author={Song, Jiaming and Meng, Chenlin and Ermon, Stefano},
  journal={arXiv preprint arXiv:2010.02502},
  year={2020}
}

@article{podell2023sdxl,
  title={Sdxl: Improving latent diffusion models for high-resolution image synthesis},
  author={Podell, Dustin and English, Zion and Lacey, Kyle and Blattmann, Andreas and Dockhorn, Tim and M{\"u}ller, Jonas and Penna, Joe and Rombach, Robin},
  journal={arXiv preprint arXiv:2307.01952},
  year={2023}
}

@article{zhong2025outdreamer,
  title={OutDreamer: Video Outpainting with a Diffusion Transformer},
  author={Zhong, Linhao and Li, Fan and Huang, Yi and Liu, Jianzhuang and Pei, Renjing and Song, Fenglong},
  journal={arXiv preprint arXiv:2506.22298},
  year={2025}
}

@article{dLLM-survey,
  title={Discrete Diffusion in Large Language and Multimodal Models: A Survey},
  author={Yu, Runpeng and Li, Qi and Wang, Xinchao},
  journal={arXiv preprint arXiv:2506.13759},
  year={2025}
}

@article{d3pm,
  title={Structured denoising diffusion models in discrete state-spaces},
  author={Austin, Jacob and Johnson, Daniel D and Ho, Jonathan and Tarlow, Daniel and Van Den Berg, Rianne},
  journal={Advances in neural information processing systems},
  volume={34},
  pages={17981--17993},
  year={2021}
}

@article{rdm,
  title={A reparameterized discrete diffusion model for text generation},
  author={Zheng, Lin and Yuan, Jianbo and Yu, Lei and Kong, Lingpeng},
  journal={arXiv preprint arXiv:2302.05737},
  year={2023}
}

@article{mdlm,
  title={Simple and effective masked diffusion language models},
  author={Sahoo, Subham and Arriola, Marianne and Schiff, Yair and Gokaslan, Aaron and Marroquin, Edgar and Chiu, Justin and Rush, Alexander and Kuleshov, Volodymyr},
  journal={Advances in Neural Information Processing Systems},
  volume={37},
  pages={130136--130184},
  year={2024}
}

@article{md4,
  title={Simplified and generalized masked diffusion for discrete data},
  author={Shi, Jiaxin and Han, Kehang and Wang, Zhe and Doucet, Arnaud and Titsias, Michalis},
  journal={Advances in neural information processing systems},
  volume={37},
  pages={103131--103167},
  year={2024}
}

@article{diffusion-llms,
  title={Diffusion language models can perform many tasks with scaling and instruction-finetuning},
  author={Ye, Jiasheng and Zheng, Zaixiang and Bao, Yu and Qian, Lihua and Gu, Quanquan},
  journal={arXiv preprint arXiv:2308.12219},
  year={2023}
}

@article{DiffuGPT-DiffuLLaMA,
  title={Scaling diffusion language models via adaptation from autoregressive models},
  author={Gong, Shansan and Agarwal, Shivam and Zhang, Yizhe and Ye, Jiacheng and Zheng, Lin and Li, Mukai and An, Chenxin and Zhao, Peilin and Bi, Wei and Han, Jiawei and others},
  journal={arXiv preprint arXiv:2410.17891},
  year={2024}
}

@article{tess-2,
  title={Tess 2: A large-scale generalist diffusion language model},
  author={Tae, Jaesung and Ivison, Hamish and Kumar, Sachin and Cohan, Arman},
  journal={arXiv preprint arXiv:2502.13917},
  year={2025}
}

@article{uncertainty-benchmarking,
  title={Benchmarking uncertainty quantification methods for large language models with lm-polygraph},
  author={Vashurin, Roman and Fadeeva, Ekaterina and Vazhentsev, Artem and Rvanova, Lyudmila and Tsvigun, Akim and Vasilev, Daniil and Xing, Rui and Sadallah, Abdelrahman Boda and Grishchenkov, Kirill and Petrakov, Sergey and others},
  journal={arXiv preprint arXiv:2406.15627},
  year={2024}
}

@article{kuhn2023semantic,
  title={Semantic uncertainty: Linguistic invariances for uncertainty estimation in natural language generation},
  author={Kuhn, Lorenz and Gal, Yarin and Farquhar, Sebastian},
  journal={arXiv preprint arXiv:2302.09664},
  year={2023}
}

@article{uq-survey1,
  title={A survey on uncertainty quantification of large language models: Taxonomy, open research challenges, and future directions},
  author={Shorinwa, Ola and Mei, Zhiting and Lidard, Justin and Ren, Allen Z and Majumdar, Anirudha},
  journal={ACM Computing Surveys},
  year={2025},
  publisher={ACM New York, NY}
}

@article{uq-survey2,
  title={A survey on uncertainty quantification methods for deep learning},
  author={He, Wenchong and Jiang, Zhe and Xiao, Tingsong and Xu, Zelin and Li, Yukun},
  journal={arXiv preprint arXiv:2302.13425},
  year={2023}
}

@article{fadeeva2024fact,
  title={Fact-checking the output of large language models via token-level uncertainty quantification},
  author={Fadeeva, Ekaterina and Rubashevskii, Aleksandr and Shelmanov, Artem and Petrakov, Sergey and Li, Haonan and Mubarak, Hamdy and Tsymbalov, Evgenii and Kuzmin, Gleb and Panchenko, Alexander and Baldwin, Timothy and others},
  journal={arXiv preprint arXiv:2403.04696},
  year={2024}
}

@inproceedings{bakman2024mars,
  title={Mars: Meaning-aware response scoring for uncertainty estimation in generative llms},
  author={Bakman, Yavuz Faruk and Yaldiz, Duygu Nur and Buyukates, Baturalp and Tao, Chenyang and Dimitriadis, Dimitrios and Avestimehr, Salman},
  booktitle={Proceedings of the 62nd Annual Meeting of the Association for Computational Linguistics (Volume 1: Long Papers)},
  pages={7752--7767},
  year={2024}
}

@article{stengel2024lacie,
  title={LACIE: Listener-aware finetuning for calibration in large language models},
  author={Stengel-Eskin, Elias and Hase, Peter and Bansal, Mohit},
  journal={Advances in Neural Information Processing Systems},
  volume={37},
  pages={43080--43106},
  year={2024}
}

@article{xu2024sayself,
  title={Sayself: Teaching llms to express confidence with self-reflective rationales},
  author={Xu, Tianyang and Wu, Shujin and Diao, Shizhe and Liu, Xiaoze and Wang, Xingyao and Chen, Yangyi and Gao, Jing},
  journal={arXiv preprint arXiv:2405.20974},
  year={2024}
}

@article{lin2022teaching,
  title={Teaching models to express their uncertainty in words},
  author={Lin, Stephanie and Hilton, Jacob and Evans, Owain},
  journal={arXiv preprint arXiv:2205.14334},
  year={2022}
}

@article{geng2023survey,
  title={A survey of confidence estimation and calibration in large language models},
  author={Geng, Jiahui and Cai, Fengyu and Wang, Yuxia and Koeppl, Heinz and Nakov, Preslav and Gurevych, Iryna},
  journal={arXiv preprint arXiv:2311.08298},
  year={2023}
}

@article{huang2024calibrating,
  title={Calibrating long-form generations from large language models},
  author={Huang, Yukun and Liu, Yixin and Thirukovalluru, Raghuveer and Cohan, Arman and Dhingra, Bhuwan},
  journal={arXiv preprint arXiv:2402.06544},
  year={2024}
}

@article{xie2024calibrating,
  title={Calibrating language models with adaptive temperature scaling},
  author={Xie, Johnathan and Chen, Annie S and Lee, Yoonho and Mitchell, Eric and Finn, Chelsea},
  journal={arXiv preprint arXiv:2409.19817},
  year={2024}
}

@inproceedings{ren2023self,
  title={Self-evaluation improves selective generation in large language models},
  author={Ren, Jie and Zhao, Yao and Vu, Tu and Liu, Peter J and Lakshminarayanan, Balaji},
  booktitle={Proceedings on},
  pages={49--64},
  year={2023},
  organization={PMLR}
}

@article{chang2024survey,
  title={A survey on evaluation of large language models},
  author={Chang, Yupeng and Wang, Xu and Wang, Jindong and Wu, Yuan and Yang, Linyi and Zhu, Kaijie and Chen, Hao and Yi, Xiaoyuan and Wang, Cunxiang and Wang, Yidong and others},
  journal={ACM transactions on intelligent systems and technology},
  volume={15},
  number={3},
  pages={1--45},
  year={2024},
  publisher={ACM New York, NY}
}

@article{DAEDAL,
  title={Beyond fixed: Training-free variable-length denoising for diffusion large language models},
  author={Li, Jinsong and Dong, Xiaoyi and Zang, Yuhang and Cao, Yuhang and Wang, Jiaqi and Lin, Dahua},
  journal={arXiv preprint arXiv:2508.00819},
  year={2025}
}

\clearpage
\appendix
\renewcommand\thesection{\Alph{section}}
\renewcommand\thefigure{S\arabic{figure}}
\renewcommand\thetable{S\arabic{table}}
\renewcommand\theequation{S\arabic{equation}}
\renewcommand\thealgorithm{S\arabic{algorithm}}

\renewcommand\theHfigure{S\arabic{figure}}
\renewcommand\theHtable{S\arabic{table}}
\renewcommand\theHequation{S\arabic{equation}}
\renewcommand\theHalgorithm{S\arabic{algorithm}}

\setcounter{figure}{0}
\setcounter{table}{0}
\setcounter{equation}{0}
\setcounter{algorithm}{0}



\twocolumn[

\section*{Appendix}

\section{Appendix Overview}

\begin{figure*}[t]
    \subsection{Examples for Comparison of Three Distributions}
    \label{sec:comparison-of-three-distributions-more-examples}
    
    As discussed in Appendix~\ref{sec:boxplot-for-distribution-distances}, we obtain the distributions corresponding to the GT token, the mask token, and a random token. Here, we directly compare these three distributions and sample multiple examples for a clearer illustration, as shown in Figure~\ref{fig:three-distribution-for-single-token-many-examples}. The results further demonstrate the effectiveness of the token regeneration approach.
    
    \centering
    \includegraphics[width=\linewidth]{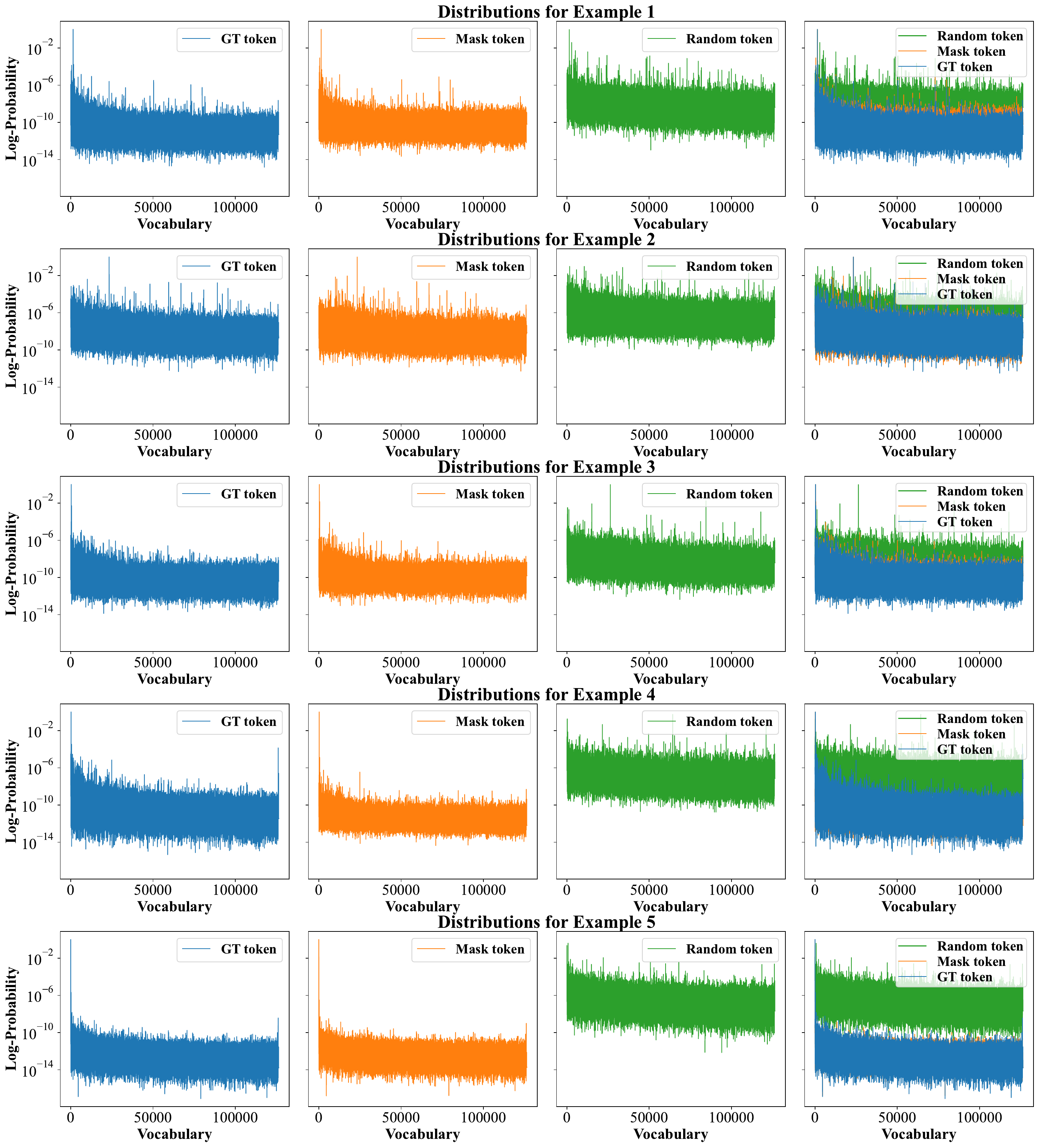}
    \caption{More examples for distributions of GT, mask, and random tokens and their comparison.}
    \label{fig:three-distribution-for-single-token-many-examples}
\end{figure*}

This appendix provides more probability estimation formulas, more details, more experimental results and more analyses to supplement the main paper. It is organized as follows:

\begin{itemize}
    \item \textbf{Appendix~\ref{sec:more-probability-estimation-formulas}: More Probability Estimation Formulas}
    \begin{itemize}
        \item Appendix~\ref{sec:unconditional-probability-estimation-arllm}: Auto-regressive LLM Probability Estimation
        \item Appendix~\ref{sec:conditional-probability-estimation-arllm}: Auto-regressive LLM Probability Estimation for Conditional Generation
        \item Appendix~\ref{sec:conditional-probability-estimation-dllm-monte-carlo}: dLLM Monte Carlo Probability Estimation for Conditional Generation
    \end{itemize}

    \item \textbf{Appendix~\ref{sec:more-details}: More Details}
    \begin{itemize}
        \item Appendix~\ref{sec:flexible-generation-algorithm}: Algorithm for Flexible-length dLLM Generation with DiSE
        \item Appendix~\ref{sec:more-implementation-details}: More Implementation Details
    \end{itemize}

    \item \textbf{Appendix~\ref{sec:more-experimental-results}: More Experimental Results}
    \begin{itemize}
        \item Appendix~\ref{sec:best-of-n-results}: Best-of-N Sampling Results
        \item Appendix~\ref{sec:additional-example-uncertainty-quantification}: Additional Qualitative Examples of Uncertainty Quantification
        \item Appendix~\ref{sec:selection-mode-ablation-more}: Ablation Study for Different Selection Modes in Uncertainty Quantification
        \item Appendix~\ref{sec:flexible-generation-ablation-more}: Ablation Study for Flexible-length dLLM Generation
        \item Appendix~\ref{sec:comparison-with-daedal}: Comparison with DAEDAL (Another Training-Free Flexible-Length dLLM Generation Method)
        \item Appendix~\ref{sec:additional-results-using-dream7B}: Additional Results using Dream-Instruct-7B
    \end{itemize}
    
    \item \textbf{Appendix~\ref{sec:more-analyses}: More analyses}
    \begin{itemize}
        \item Appendix~\ref{sec:rank-distribution-statistics-more}: Rank Distribution Statistics for Fixed Sequences and Positions
        \item Appendix~\ref{sec:boxplot-for-distribution-distances}: Boxplot for Distribution Distances
        \item Appendix~\ref{sec:comparison-of-three-distributions-more-examples}: Examples for Comparison of Three Distributions
    \end{itemize}

\end{itemize}

]

\begin{algorithm*}[ht]
\caption{Flexible-length dLLM Generation with DiSE}
\label{alg:flexible}
\begin{algorithmic}[1]
\Require Prompt $P$, base length $L$, maximum iterations $M_{max}$, patience $K$, mask size $D$
\Ensure Final generated sequence $\hat{X}$
\State Generate an initial response $R$ of length $L$ given prompt $P$
\State Remove all EOT tokens from $R$ to obtain $\overline{R}$
\State $X^{(1)} = [P; \overline{R}]$
\State Compute initial confidence $s^{(1)} \gets \mathrm{DiSE}(X^{(1)})$
\State Set $\hat{X} \gets X^{(1)}$, $\hat{s} \gets s^{(1)}$, $t \gets 1$, $c \gets 0$
\While{$t < M_{max}$}
    \State $t \gets t + 1$
    \State Mask the last $D$ tokens of $X^{(t-1)}$ to obtain $X_m^{(t-1)}$
    \State Regenerate sequence $X^{(t)}$ from the masked input $[X_m^{(t-1)}; \langle \mathrm{mask\ token} \rangle]$
    \State $s^{(t)} \gets \mathrm{DiSE}(X^{(t)})$
    \If{$s^{(t)} > \hat{s}$}
        \State $\hat{X} \gets X^{(t)}$, $\hat{s} \gets s^{(t)}$, $c \gets 0$
    \Else
        \State $c \gets c + 1$
    \EndIf
    \If{$c \geq K$}
        \State \textbf{break}
    \EndIf
    \State $D \gets D + 1$
\EndWhile
\State \Return $\hat{X}$
\end{algorithmic}
\end{algorithm*}

\section{More Probability Estimation Formulas}
\label{sec:more-probability-estimation-formulas}

\subsection{Auto-regressive LLM Probability Estimation}
\label{sec:unconditional-probability-estimation-arllm}

Given an auto-regressive language model and a text sequence $X = (x_1, x_2, \dots, x_N)$, the probability of generating the entire sequence is factorized as the product of conditional probabilities:
\begin{equation}
    p_\theta(X) = \prod_{i=1}^{N} p_\theta(x_i \mid x_{<i}),
\end{equation}
where $x_{<i} = (x_1, \dots, x_{i-1})$ represents all preceding tokens, and $\theta$ denotes the model parameters. This factorization allows exact computation of the sequence probability by multiplying the model's predicted probabilities for each token given its context. 

\subsection{Auto-regressive LLM Probability Estimation for Conditional Generation}
\label{sec:conditional-probability-estimation-arllm}

In the context of conditional generation given a prompt $P$, let $R = (r_1, r_2, \dots, r_N)$ denote the generated response of length $N$. The probability of generating $R$ given $P$ for an auto-regressive language model can be written as:
\begin{equation}
    p_\theta(R \mid P) = \prod_{i=1}^{N} p_\theta(r_i \mid P, r_{<i}),
\end{equation}
where $r_{<i} = (r_1, \dots, r_{i-1})$. This formulation allows exact computation of the probability of a model-generated response conditioned on a given prompt.

\subsection{dLLM Monte Carlo Probability Estimation for Conditional Generation}
\label{sec:conditional-probability-estimation-dllm-monte-carlo}


For dLLMs, let $R^0 = (r^0_1, r^0_2, \dots, r^0_N)$ denote the generated response of length $N$. The traditional dLLM approach approximates the log-probability of generating $R^0$ given $P$ with the following term:

\begin{small}
\begin{equation}
\mathbb{E}_{l, R^l}\left[\frac{N}{l} \sum_{i=1}^{N} \mathbf{1}\left[r^l_i= \langle \mathrm{mask\ token} \rangle \right] \log p_\theta\left(r^0_i \mid P, R^l\right)\right],
\end{equation}
\end{small}
where $l$ is uniformly sampled from $\{1, 2, \dots, N\}$, and $R^l=(r^l_1, r^l_2, \dots, r^l_N)$ is obtained by uniformly sampling $l$ tokens from $R^0$, replacing the tokens at these positions with mask tokens, while keeping all other tokens identical to those in $X^0$.
Since the exact computation of this expectation is intractable, we employ Monte Carlo simulation to approximate it by sampling a finite number of instances and taking their empirical average.

\section{More Details}
\label{sec:more-details}

\subsection{Algorithm for Flexible-length dLLM Generation with DiSE}
\label{sec:flexible-generation-algorithm}

We provide a detailed algorithm for the flexible-length dLLM generation framework guided by the DiSE score in Algorithm~\ref{alg:flexible}, which uses the DiSE score as a self-evaluation signal to achieve controllable sequence lengths and improved generation quality.

\begin{table*}[t] 
\centering
\caption{Best-of-N sampling results for uncertainty quantification on the Countdown, GSM8K, MATH500, and SVAMP datasets with varying generation lengths. The table compares the baseline without best-of-N sampling, Monte Carlo simulation with varying $N_{mc}$, the proposed DiSE, and the perplexity calculation using the auto-regressive model LLaMA3-Instruct-8B. The last column reports the average accuracy across the preceding 12 settings.}
\label{tab:best-of-N} 
\resizebox{\textwidth}{!}{
    \begin{tabular}{ll|cccccccccccc|>{\columncolor{yellow!20}}c}
        \toprule
        &
        & \multicolumn{3}{c}{\textbf{Countdown}}
        & \multicolumn{3}{c}{\textbf{GSM8K}} 
        & \multicolumn{3}{c}{\textbf{MATH500}} 
        & \multicolumn{3}{c|}{\textbf{SVAMP}} 
        & \multirow{2}{*}{\textbf{Avg. Accuracy}} \\
        \cmidrule(lr){3-5} \cmidrule(lr){6-8} \cmidrule(lr){9-11} \cmidrule(lr){12-14}
         & \textbf{Method / Gen Len} & 128 & 256 & 512 & 128 & 256 & 512 & 128 & 256 & 512 & 128 & 256 & 512 & \\
        \midrule
        \multirow{6}{*}{\textbf{LLaDA-Instruct-8B}} 
        & Baseline   & 26.17 & 15.23 & 12.50 & 68.01 & 76.65 & 79.23 & 26.20 & 32.80 & 36.80 & 84.67 & 85.00 & 83.67 & 52.24 \\
        \cmidrule(lr){2-15}
        & MC, $N_{mc}=1$ & 24.61 & 21.48 & 17.19 & 68.84 & 78.17 & 80.59 & 25.80 & 33.80 & 36.40 & 84.67 & 84.00 & 85.00 & 53.38 \\
        \cmidrule(lr){2-15}
        & MC, $N_{mc}=32$ & 29.69 & 21.88 & 16.41 & 71.11 & 78.70 & 82.79 & 27.60 & \textbf{34.80} & 36.20 & 86.33 & 85.67 & 86.67 & 54.82 \\
        \cmidrule(lr){2-15}

        
        & \textbf{DiSE (ours)} & \textbf{30.86} & \textbf{24.22} & \textbf{27.34} & \textbf{73.01} & \textbf{82.41} & \textbf{83.02} & \textbf{29.80} & 34.60 & \textbf{38.20} & \textbf{88.33} & \textbf{87.00} & \textbf{90.00} & \textbf{57.40} \\

        \cmidrule(lr){2-15}
        \morecmidrules
        \cmidrule(lr){2-15}
        & \textcolor{gray}{LLaMA perplexity} & \textcolor{gray}{30.86} & \textcolor{gray}{17.19} & \textcolor{gray}{11.33} & \textcolor{gray}{74.22} & \textcolor{gray}{79.61} & \textcolor{gray}{81.20} & \textcolor{gray}{28.60} & \textcolor{gray}{35.40} & \textcolor{gray}{34.80} & \textcolor{gray}{88.33} & \textcolor{gray}{86.67} & \textcolor{gray}{85.67} & \textcolor{gray}{54.49} \\
        \midrule
        \midrule
        \multirow{6}{*}{\textbf{LLaDA-1.5-8B}} 
        & Baseline   & 24.22 & 15.62 & 17.19 & 70.51 & 77.48 & 79.53 & 26.80 & 34.00 & 36.80 & 87.00 & 84.67 & 86.67 & 53.37\\
        \cmidrule(lr){2-15}
        & MC, $N_{mc}=1$ & 24.22 & 20.31 & 24.22 & 70.13 & 79.45 & 80.89 & 28.20 & 34.80 & 37.60 & \textbf{88.33} & 84.67 & 87.33 & 55.01\\
        \cmidrule(lr){2-15}
        & MC, $N_{mc}=32$ & 26.17 & \textbf{20.70} & 21.88 & 72.63 & 79.91 & 82.79 & 28.40 & \textbf{35.60} & \textbf{38.80} & 88.00 & 85.33 & 86.33 & 55.55\\
        \cmidrule(lr){2-15}
        
        & \textbf{DiSE (ours)} & \textbf{29.30} & 17.97 & \textbf{28.91} & \textbf{74.53} & \textbf{81.96} & \textbf{83.55} & \textbf{28.60} & 34.40 & 37.40 & 88.00 & \textbf{86.33} & \textbf{87.67} & \textbf{56.55} \\

        \cmidrule(lr){2-15}
        \morecmidrules
        \cmidrule(lr){2-15}
        & \textcolor{gray}{LLaMA perplexity} & \textcolor{gray}{28.91} & \textcolor{gray}{12.89} & \textcolor{gray}{13.28} & \textcolor{gray}{74.60} & \textcolor{gray}{81.50} & \textcolor{gray}{80.14} & \textcolor{gray}{30.40} & \textcolor{gray}{39.00} & \textcolor{gray}{37.20} & \textcolor{gray}{89.67} & \textcolor{gray}{87.67} & \textcolor{gray}{87.00} & \textcolor{gray}{55.19} \\
        \bottomrule
    \end{tabular}
}
\end{table*}

\subsection{More Implementation Details}
\label{sec:more-implementation-details}

We evaluate the performance of our model across the following benchmarks.

\begin{itemize}
    \item \textbf{ARC-Challenge}~\citep{dataset-arc-challenge}: A subset of 7,787 grade-school science questions specifically filtered to exclude those solvable by simple retrieval or co-occurrence methods, thereby necessitating advanced logical reasoning.
    \item \textbf{GPQA}~\citep{dataset-gpqa}: A highly rigorous benchmark of 448 expert-written multiple-choice questions in biology, physics, and chemistry. These ``Google-proof'' problems are designed to challenge even domain-specific PhDs and evaluate scalable oversight for AI systems that surpass human-level performance.
    \item \textbf{Countdown}~\citep{dataset-countdown}: A combinatorial task where models must derive a target integer from a provided set of numbers using elementary arithmetic operations.
    \item \textbf{GSM8K}~\citep{dataset-gsm8k}: A benchmark comprising 8.5K grade-school level math problems that necessitate 2--8 steps of multi-step logical reasoning.
    \item \textbf{MATH500}~\citep{dataset-math500}: A curated subset of 500 high-school competition problems from the MATH dataset, targeting advanced problem-solving capabilities.
    \item \textbf{SVAMP}~\citep{dataset-svamp}: A collection of 1K elementary word problems specifically designed to evaluate model robustness against diverse linguistic framing and narrative variations.
\end{itemize}

The datasets employed in our experiments are categorized into two groups: those used for conditional likelihood estimation and those intended for conditional generation.
Specifically, we consider ARC-Challenge~\citep{dataset-arc-challenge} and GPQA~\citep{dataset-gpqa} for conditional likelihood estimation, which are challenging multiple-choice science question datasets.
ARC-Challenge focuses on grade-school level questions that require advanced reasoning beyond simple retrieval, while GPQA contains expert-crafted questions in biology, physics, and chemistry that are difficult even for highly skilled humans and state-of-the-art AI models. 
The generation process is configured to produce two tokens per step. All experiments employ a semi-autoregressive decoding strategy~\citep{llada} with a block size of $32$.
For conditional generation, we use Countdown~\citep{dataset-countdown}, GSM8K~\citep{dataset-gsm8k}, MATH500~\citep{dataset-math500}, and SVAMP~\citep{dataset-svamp}, which involve arithmetic and mathematical problems requiring step-by-step reasoning, advanced problem-solving, combinatorial thinking, and generalization across diverse problem formats.
Regarding the selection mode, i.e., the binary mask $M$, we adopt different configurations for different datasets. For ARC-Challenge, we set $M=1$ for the last two tokens of the prompt $P$. For GPQA, we set $M=1$ for the last seven tokens of the prompt $P$ and the first two tokens of the response $R$. For Countdown, GSM8K, MATH500 and SVAMP, we adopt the selection mode `last-10' by default, which sets $M=1$ only for the last ten non-EOT tokens.
For flexible-length dLLM generation experiments, we set the maximum number of iterations $M_{max}=10$, the patience parameter $K=4$, and the mask size $D=20$ by default.

\begin{figure*}[t]
    \centering
    \includegraphics[width=\linewidth]{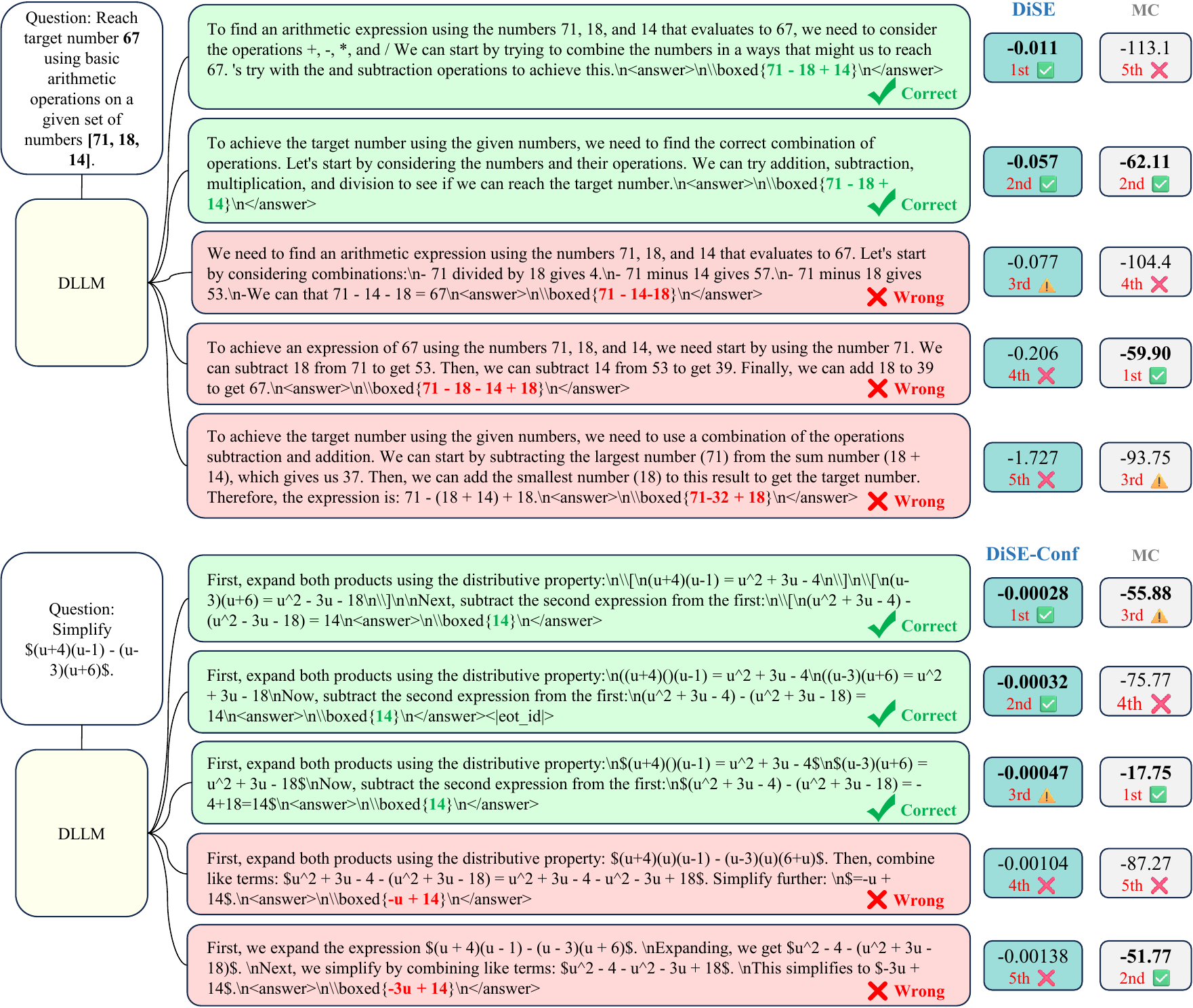}
    \caption{Additional qualitative examples of uncertainty quantification using LLaDA-Instruct-8B. DiSE assigns higher scores to correct answers and lower scores to incorrect answers, while the Monte Carlo simulation ($N_{mc}=32$) produces scores that do not consistently reflect correctness.}
    \label{fig:qualitative-results-additional}
\end{figure*}

\begin{table*}[t] 
\centering
\caption{Additional ROC-AUC results for uncertainty quantification to investigate the impact of different selection modes on performance. We evaluate two selection modes `full' and `last-10'. The table reports ROC-AUC scores across the Countdown, GSM8K, MATH500, and SVAMP datasets with varying generation lengths, as well as the average ROC-AUC scores over all settings.}
\label{tab:ROC-AUC-more} 
\resizebox{\textwidth}{!}{
    \begin{tabular}{ll|cccccccccccc|>{\columncolor{yellow!20}}c}
        \toprule
        &
        & \multicolumn{3}{c}{\textbf{Countdown}}
        & \multicolumn{3}{c}{\textbf{GSM8K}} 
        & \multicolumn{3}{c}{\textbf{MATH500}} 
        & \multicolumn{3}{c|}{\textbf{SVAMP}} 
        & \multirow{2}{*}{\textbf{Avg. ROC-AUC$\uparrow$}} \\
        \cmidrule(lr){3-5} \cmidrule(lr){6-8} \cmidrule(lr){9-11} \cmidrule(lr){12-14}
         & \textbf{Method / Gen Len} & 128 & 256 & 512 & 128 & 256 & 512 & 128 & 256 & 512 & 128 & 256 & 512 & \\
        \midrule
        \multirow{2}{*}{\textbf{LLaDA-Instruct-8B}} 
        & \textbf{DiSE (full)} & 0.616 & 0.672 & 0.698 & 0.597 & 0.585 & 0.560 & 0.514 & 0.555 & 0.517 & 0.665 & 0.549 & 0.571 & 0.592 \\
        \cmidrule(lr){2-15}
        & \textbf{DiSE (last-10)} & 0.578 & 0.521 & 0.622 & 0.633 & 0.644 & 0.658 & 0.611 & 0.634 & 0.604 & 0.688 & 0.692 & 0.755 & 0.637 \\

        \midrule
        \midrule
        \multirow{2}{*}{\textbf{LLaDA-1.5-8B}} 
        & \textbf{DiSE (full)} & 0.591 & 0.664 & 0.681 & 0.593 & 0.569 & 0.546 & 0.489 & 0.590 & 0.532 & 0.630 & 0.574 & 0.552 & 0.584\\
        \cmidrule(lr){2-15}
        & \textbf{DiSE (last-10)} & 0.610 & 0.471 & 0.586 & 0.610 & 0.616 & 0.613 & 0.606 & 0.553 & 0.533 & 0.599 & 0.629 & 0.677 & 0.592 \\
        \bottomrule
    \end{tabular}
}
\end{table*}

\begin{table*}[t] 
\centering
\caption{Additional best-of-N sampling results for uncertainty quantification to investigate the impact of different selection modes on performance. We evaluate two selection modes `full' and `last-10'. The table reports accuracy across the Countdown, GSM8K, MATH500, and SVAMP datasets with varying generation lengths, as well as the average accuracy over all settings.}
\label{tab:best-of-N-more} 
\resizebox{\textwidth}{!}{
    \begin{tabular}{ll|cccccccccccc|>{\columncolor{yellow!20}}c}
        \toprule
        &
        & \multicolumn{3}{c}{\textbf{Countdown}}
        & \multicolumn{3}{c}{\textbf{GSM8K}} 
        & \multicolumn{3}{c}{\textbf{MATH500}} 
        & \multicolumn{3}{c|}{\textbf{SVAMP}} 
        & \multirow{2}{*}{\textbf{Avg. Accuracy}} \\
        \cmidrule(lr){3-5} \cmidrule(lr){6-8} \cmidrule(lr){9-11} \cmidrule(lr){12-14}
         & \textbf{Method / Gen Len} & 128 & 256 & 512 & 128 & 256 & 512 & 128 & 256 & 512 & 128 & 256 & 512 & \\
        \midrule
        \multirow{2}{*}{\textbf{LLaDA-Instruct-8B}} 
        & \textbf{DiSE (full)} & 30.86 & 28.52 & 27.34 & 71.87 & 79.76 & 79.53 & 27.20 & 34.60 & 34.20 & 87.67 & 85.33 & 87.00 & 56.16 \\
        \cmidrule(lr){2-15}
        & \textbf{DiSE (last-10)} & 30.86 & 24.22 & 27.34 & 73.01 & 82.41 & 83.02 & 29.80 & 34.60 & 38.20 & 88.33 & 87.00 & 90.00 & 57.40 \\

        \midrule
        \midrule
        \multirow{2}{*}{\textbf{LLaDA-1.5-8B}} 
        & \textbf{DiSE (full)} & 27.34 & 25.00 & 32.81 & 72.33 & 79.45 & 80.06 & 24.80 & 37.20 & 38.00 & 88.33 & 86.67 & 85.00 & 56.42 \\
        \cmidrule(lr){2-15}
        & \textbf{DiSE (last-10)} & 29.30 & 17.97 & 28.91 & 74.53 & 81.96 & 83.55 & 28.60 & 34.40 & 37.40 & 88.00 & 86.33 & 87.67 & 56.55 \\

        \bottomrule
    \end{tabular}
}
\end{table*}

\section{More Experimental Results}
\label{sec:more-experimental-results}

\subsection{Best-of-N Sampling Results}
\label{sec:best-of-n-results}

In Section~\ref{sec:experiments-uncertainty-quantification}, we generate multiple answers for each question and evaluate uncertainty quantification using ROC-AUC scores. As an additional experiment, we perform best-of-N sampling, selecting the answer with the lowest uncertainty (i.e., highest DiSE score in our proposed method) among multiple generations per question, and report the accuracy. Consistent with the main experiments, we generate five answers per question.
Table~\ref{tab:best-of-N} presents the evaluation results of our method under the best-of-N sampling strategy on the Countdown, GSM8K, MATH500, and SVAMP datasets with varying generation lengths. The results demonstrate that our approach consistently outperforms the baseline method that does not employ best-of-N sampling across all tested configurations, highlighting the effectiveness of selecting the highest-scoring candidate based on DiSE. In comparison to the conventional Monte Carlo simulation method, our approach yields substantially larger improvements. In particular, when using the LLaDA-Instruct-8B model, the proposed method achieves an average accuracy gain of $5.16\%$ over all twelve generation length settings, whereas the Monte Carlo method with a comparable computational cost, corresponding to $N_{mc}=1$, achieves only an improvement of $1.14\%$. Even when the Monte Carlo method is applied with $N_{mc}=32$, resulting in an evaluation cost nearly $32$ times higher, the observed improvement reaches only $2.58\%$, which is still considerably lower than the gain provided by our approach. Furthermore, we evaluate performance using probability estimates obtained from an auto-regressive LLM as a reference. For instance, under the same generations, employing the auto-regressive LLM probabilities leads to an improvement of merely $2.25\%$, which remains below the performance enhancement achieved by our method, thereby underscoring the superiority of DiSE in best-of-N sampling and uncertainty quantification.
Importantly, the observed improvements are consistent across both tested dLLM variants, LLaDA-Instruct-8B and LLaDA-1.5-8B, across four datasets and three generation lengths. This consistency indicates that best-of-N sampling guided by DiSE remains robust regardless of model, task type or sequence length. 

\subsection{Additional Qualitative Examples of Uncertainty Quantification}
\label{sec:additional-example-uncertainty-quantification}

Figure~\ref{fig:qualitative-results-additional} presents additional qualitative examples of uncertainty quantification using LLaDA-Instruct-8B. 
Consistently, DiSE effectively distinguishes between correct and incorrect outputs by assigning higher scores to correct answers, corresponding to lower uncertainty, while the Monte Carlo simulation with $N_{mc}=32$ fails to align with the correctness of the answers.
These results provide additional evidence of the effectiveness of DiSE as a fine-grained uncertainty measure at the sequence level.

\subsection{Ablation Study for Different Selection Modes in Uncertainty Quantification}
\label{sec:selection-mode-ablation-more}

In Section~\ref{sec:experiments-uncertainty-quantification}, we report the effectiveness of DiSE for uncertainty quantification under the selection mode `last-10', showing substantial improvements over the baseline across multiple datasets and generation lengths. To further validate the robustness of this finding, we extend the analysis by additionally evaluating the selection mode `full' configuration and directly comparing it with the mode `last-10'. The ROC-AUC results are presented in Table~\ref{tab:ROC-AUC-more} and the best-of-N sampling results are presented in Table~\ref{tab:best-of-N-more}. 
Without specifying a local region for computing regeneration probability, DiSE with `full' mode still achieves performance far above the baseline, demonstrating the effectiveness of our method.

\begin{table*}[ht] 
\centering
\caption{Ablation study on flexible-length dLLM generation using LLaDA-Instruct-8B, analyzing the impact of different patience values $K$ on performance. The table reports accuracy on the Countdown, GSM8K, MATH500, and SVAMP datasets with varying base lengths, along with the average accuracy, the average number of model forward passes and average time for each configuration.}
\label{tab:flexible-length-generation-ablation-llada-instruct} 
\resizebox{\textwidth}{!}{
    \begin{tabular}{l|cccccccccccc|>{\columncolor{yellow!20}}c|>{\columncolor{gray!20}}c|>{\columncolor{gray!20}}c}
        \toprule
        & \multicolumn{3}{c}{\textbf{Countdown}}
        & \multicolumn{3}{c}{\textbf{GSM8K}} 
        & \multicolumn{3}{c}{\textbf{MATH500}} 
        & \multicolumn{3}{c|}{\textbf{SVAMP}} 
        & \multirow{2}{*}{\textbf{Avg. Accuracy}}
        & \multirow{2}{*}{\textbf{Avg. \# NFE}}
        & \multirow{2}{*}{\textbf{Avg. Time(s)}} \\
        \cmidrule(lr){2-4} \cmidrule(lr){5-7} \cmidrule(lr){8-10} \cmidrule(lr){11-13}
        \textbf{Method / Base Len} & 128 & 256 & 512 & 128 & 256 & 512 & 128 & 256 & 512 & 128 & 256 & 512 & & & \\
        \midrule
        Baseline   & 26.17 & 15.23 & 12.50 & 68.01 & 76.65 & 79.23 & 26.20 & 32.80 & 36.80 & 84.67 & 85.00 & 83.67 & 52.24 & 149.3  & 3.27 \\
        \midrule
        \textbf{DiSE-flexible (K=2)} & 27.34 & 18.36 & 16.02 & 70.43 & 79.30 & 79.23 & 26.40 & 34.00 & 36.60 & 87.33 & 86.00 & 84.67 & 53.81 & 182.3  & 3.86 \\
        \midrule
        \textbf{DiSE-flexible (K=3)} & 27.34 & 17.97 & 15.62 & 70.74 & 79.61 & 79.23 & 25.80 & 33.80 & 36.60 & 87.33 & 86.00 & 84.33 & 53.70 & 199.3  & 4.18 \\
        \midrule
        \textbf{DiSE-flexible (K=4)} & 27.73 & 18.36 & 15.62 & 70.96 & 79.68 & 79.30 & 26.00 & 33.60 & 36.60 & 87.33 & 86.00 & 84.33 & 53.79 & 215.3  & 4.55 \\

        \bottomrule
    \end{tabular}
}
\end{table*}

\begin{table*}[t] 
\centering
\caption{Ablation study on flexible-length dLLM generation using LLaDA-1.5-8B, analyzing the impact of different patience values $K$ on performance. The table reports accuracy on the Countdown, GSM8K, MATH500, and SVAMP datasets with varying base lengths, along with the average accuracy, the average number of model forward passes and average time for each configuration.}
\label{tab:flexible-length-generation-ablation-llada-1.5} 
\resizebox{\textwidth}{!}{
    \begin{tabular}{l|cccccccccccc|>{\columncolor{yellow!20}}c|>{\columncolor{gray!20}}c|>{\columncolor{gray!20}}c}
        \toprule
        & \multicolumn{3}{c}{\textbf{Countdown}}
        & \multicolumn{3}{c}{\textbf{GSM8K}} 
        & \multicolumn{3}{c}{\textbf{MATH500}} 
        & \multicolumn{3}{c|}{\textbf{SVAMP}} 
        & \multirow{2}{*}{\textbf{Avg. Accuracy}}
        & \multirow{2}{*}{\textbf{Avg. \# NFE}}
        & \multirow{2}{*}{\textbf{Avg. Time(s)}} \\
        \cmidrule(lr){2-4} \cmidrule(lr){5-7} \cmidrule(lr){8-10} \cmidrule(lr){11-13}
        \textbf{Method / Base Len} & 128 & 256 & 512 & 128 & 256 & 512 & 128 & 256 & 512 & 128 & 256 & 512 & & & \\
        \midrule
        Baseline   & 24.22 & 15.62 & 17.19 & 70.51 & 77.48 & 79.53 & 26.80 & 34.00 & 36.80 & 87.00 & 84.67 & 86.67 & 53.37 & 149.3  & 3.25 \\
        \midrule
        \textbf{DiSE-flexible (K=2)} & 24.61 & 17.19 & 19.14 & 72.40 & 79.23 & 80.06 & 27.40 & 35.60 & 37.40 & 87.33 & 84.67 & 87.00 & 54.34 & 182.0 & 3.84 \\
        \midrule
        \textbf{DiSE-flexible (K=3)} & 24.61 & 19.14 & 20.70 & 72.48 & 79.45 & 80.06 & 27.80 & 35.40 & 37.40 & 87.00 & 84.67 & 87.00 & 54.64 & 198.9 & 4.20 \\
        \midrule
        \textbf{DiSE-flexible (K=4)} & 26.17 & 19.53 & 22.27 & 72.33 & 79.53 & 80.06 & 27.20 & 35.60 & 37.40 & 87.00 & 85.00 & 87.00 & 54.92 & 214.7 & 4.53 \\

        \bottomrule
    \end{tabular}
}
\end{table*}

\subsection{Ablation Study for Flexible-length dLLM Generation}
\label{sec:flexible-generation-ablation-more}

\begin{figure}[t]
    \centering
    \includegraphics[width=\linewidth]{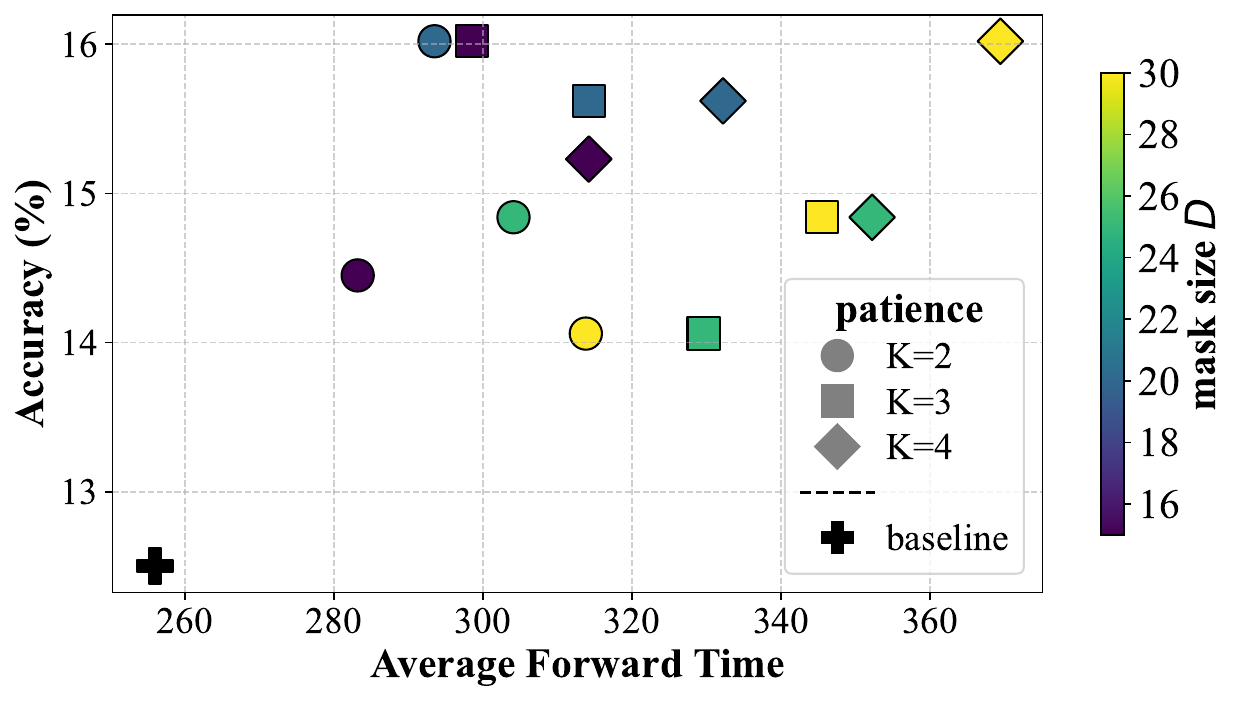}
    \caption{Ablation study on the Countdown dataset using LLaDA-Instruct-8B with base length $L=512$ for flexible-length dLLM generation with DiSE, examining the effects of different patience $K$ and mask sizes $D$ on performance. The figure presents both accuracy and the average number of model forward passes for each configuration.}
    \label{fig:flexible-ablation-llada-instruct}
\end{figure}

\begin{figure}[ht]
    \centering
    \includegraphics[width=\linewidth]{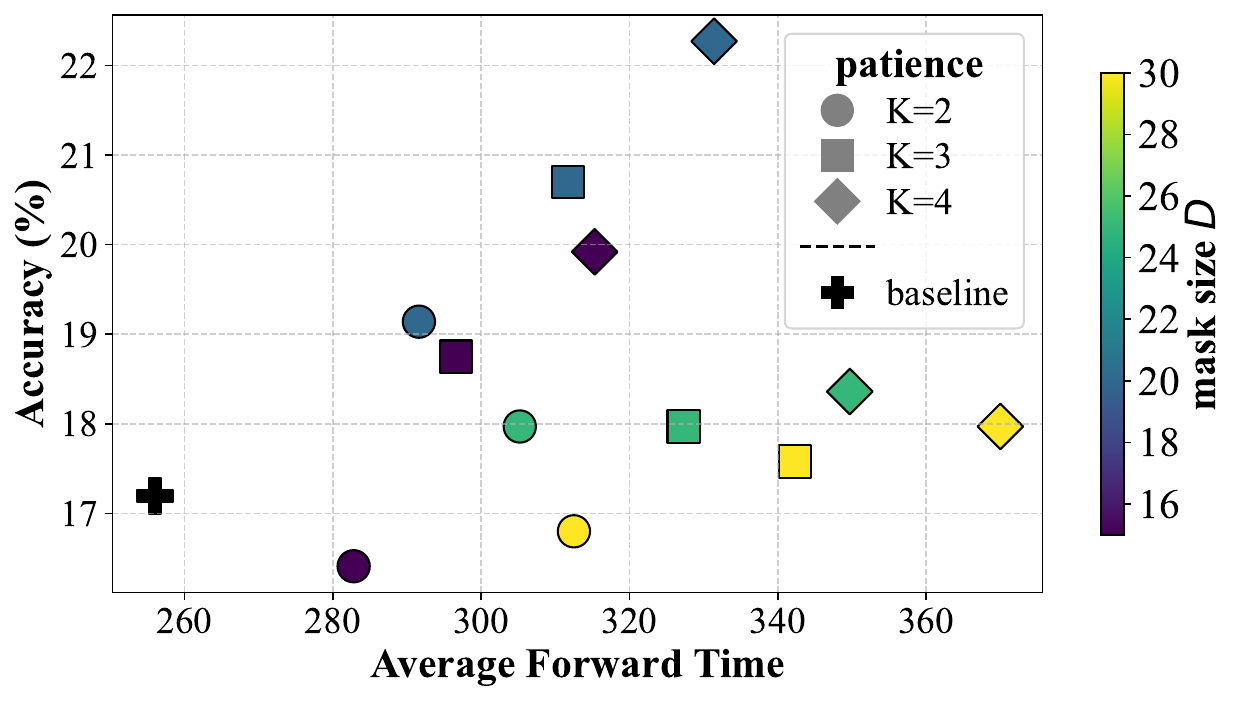}
    \caption{Ablation study on the Countdown dataset using LLaDA-1.5-8B with base length $L=512$ for flexible-length dLLM generation with DiSE, examining the effects of different patience $K$ and mask sizes $D$ on performance. The figure presents both accuracy and the average number of model forward passes for each configuration.}
    \label{fig:flexible-ablation-llada-1.5}
\end{figure}

To assess the impact of patience $K$ and mask size $D$, we perform an ablation on Countdown using LLaDA-Instruct-8B and LLaDA-1.5-8B with base length $L=512$, reporting accuracy and average forward passes in Figure~\ref{fig:flexible-ablation-llada-instruct} and Figure~\ref{fig:flexible-ablation-llada-1.5}. Our method generally outperforms the baseline, while different $K$ and $D$ settings highlight a trade-off between computational cost and performance. 


We investigate the effect of different patience values $K$ on flexible-length dLLM generation across the Countdown, GSM8K, MATH500 and SVAMP datasets with varying base lengths, testing under both the LLaDA-Instruct-8B and LLaDA-1.5-8B models. The summarized results are presented in Table~\ref{tab:flexible-length-generation-ablation-llada-instruct} and Table~\ref{tab:flexible-length-generation-ablation-llada-1.5}. 
Across all tested patience $K$ settings, the flexible-length generation guided by DiSE consistently achieves substantially better average accuracy than fixed-length baselines, demonstrating the effectiveness of adaptive sequence length. 
Increasing $K$ raises computational costs, but the corresponding performance gains are not always proportional, highlighting the need to balance efficiency with achievable improvements.

\subsection{Comparison with DAEDAL (Another Training-Free Flexible-Length dLLM Generation Method)}
\label{sec:comparison-with-daedal}

\begin{figure}[t]
    \centering
    \includegraphics[width=\linewidth]{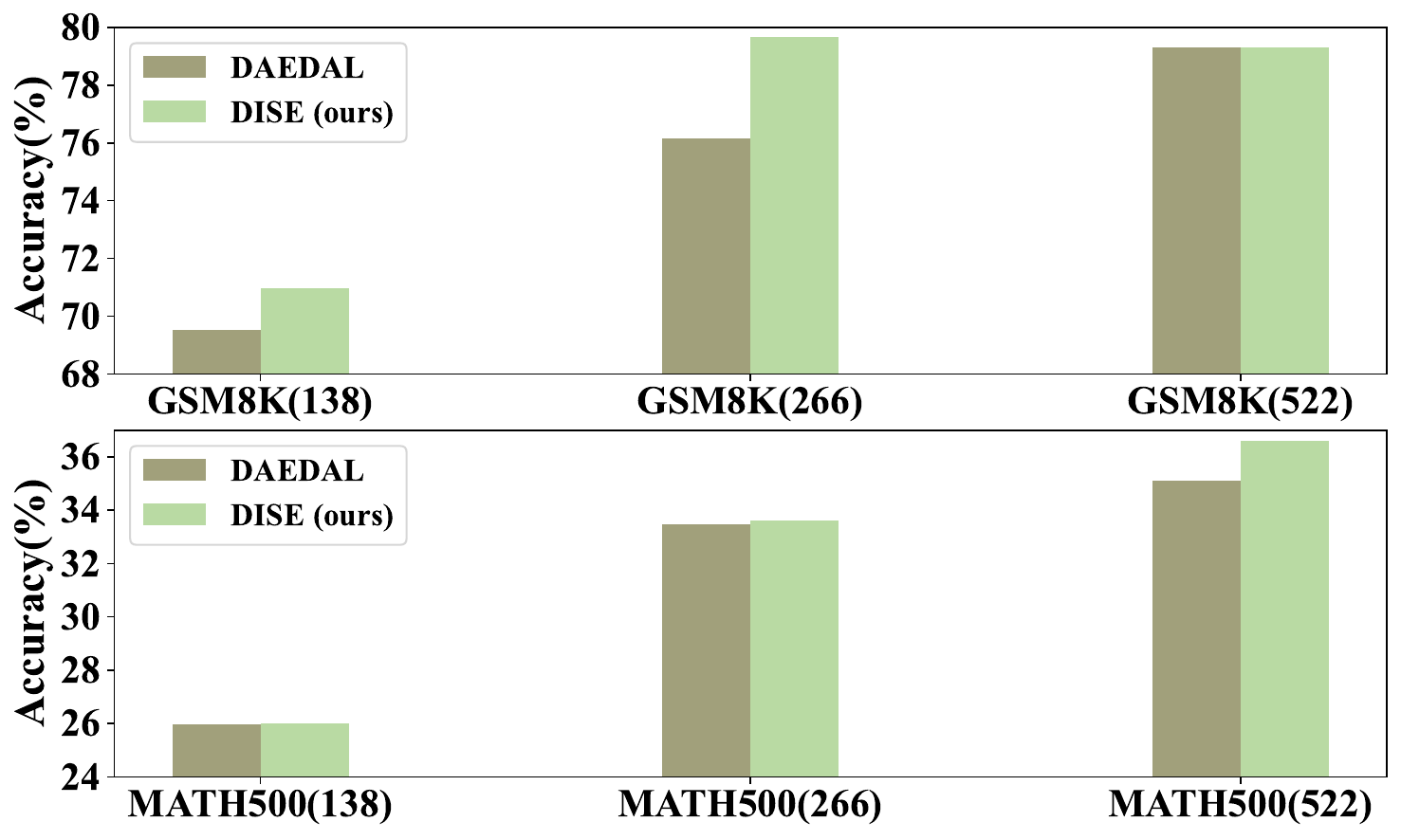}
    \caption{Comparison between our method and DAEDAL on the GSM8K and MATH500 datasets using LLaDA-Instruct-8B. The numbers in parentheses on the x-axis indicate the maximum output length for each setting.}
    \label{fig:flexible-comparison}
\end{figure}

We compare our flexible-length dLLM Generation with DiSE against DAEDAL~\citep{DAEDAL}, another training-free flexible-length dLLM approach, using LLaDA-Instruct-8B under identical experimental settings, with the same maximum output length and the same number of tokens denoised per step. As shown in Figure~\ref{fig:flexible-comparison}, our method achieves superior performance on the GSM8K and MATH500 datasets compared to DAEDAL, demonstrating both the effectiveness and stability of our approach.

\subsection{Additional Results using Dream-Instruct-7B}
\label{sec:additional-results-using-dream7B}

To further validate the generalizability of our approach, we conduct additional experiments on conditional likelihood estimation and uncertainty quantification using Dream-Instruct-7B. The results for conditional likelihood estimation are presented in Table~\ref{tab:conditional-likelihood-estimation-dream}, while those for uncertainty quantification are shown in Table~\ref{tab:ROC-AUC-dream}. Our method consistently outperforms the baseline in both tasks, further demonstrating its effectiveness and versatility. These results indicate that our approach can be easily applied to other dLLM models trained with similar methodologies.

\begin{table}[ht] 
\centering
\caption{Conditional likelihood estimation results on ARC-Challenge and GPQA using Dream-Instruct-7B. The table compares the proposed DiSE against the Monte Carlo simulation baseline with varying $N_{mc}$. The last column reports the average number of model forward passes per computation.}
\label{tab:conditional-likelihood-estimation-dream} 
\resizebox{\linewidth}{!}{
    \begin{tabular}{ll|cc|c}
        \toprule
        & \textbf{Method} & \textbf{ARC-Challenge} & \textbf{GPQA} & \textbf{\# NFE}  \\
        \midrule
        \multirow{4}{*}{\textbf{Dream-Instruct-7B}} 
        & MC, $N_{mc}=1$ & 0.290 & 0.223 & 1 \\
        \cmidrule(lr){2-5}
        & MC, $N_{mc}=32$ & 0.316 & 0.259 & 32 \\
        \cmidrule(lr){2-5}
        & \textbf{DiSE (ours)} & \textbf{0.472} & \textbf{0.297} & 1 \\
        \bottomrule
    \end{tabular}
}
\end{table}

\begin{table*}[t] 
\centering
\caption{Additional ROC-AUC results for uncertainty quantification using Dream-Instruct-7B. The table reports ROC-AUC scores across the Countdown, GSM8K, MATH500, and SVAMP datasets with varying generation lengths, as well as the average ROC-AUC scores over all settings.}
\label{tab:ROC-AUC-dream} 
\resizebox{\textwidth}{!}{
    \begin{tabular}{ll|cccccccc|>{\columncolor{yellow!20}}c}
        \toprule
        &
        & \multicolumn{2}{c}{\textbf{Countdown}}
        & \multicolumn{2}{c}{\textbf{GSM8K}} 
        & \multicolumn{2}{c}{\textbf{MATH500}} 
        & \multicolumn{2}{c|}{\textbf{SVAMP}} 
        & \multirow{2}{*}{\textbf{Avg. ROC-AUC$\uparrow$}} \\
        \cmidrule(lr){3-4} \cmidrule(lr){5-6} \cmidrule(lr){7-8} \cmidrule(lr){9-10}
         & \textbf{Method / Gen Len} & 128 & 256 & 128 & 256 & 128 & 256 & 128 & 256 & \\
        \midrule
        \multirow{3}{*}{\textbf{Dream-Instruct-7B}} 
        & MC, $N_{mc}=1$ & 0.463 & 0.548 & 0.473 & 0.503 & 0.538 & 0.464 & 0.510 & 0.551 & 0.506 \\
        \cmidrule(lr){2-11}
        & MC, $N_{mc}=32$ & 0.600 & 0.482 & 0.483 & 0.497 & 0.471 & 0.576 & 0.469 & 0.467 & 0.506 \\
        \cmidrule(lr){2-11}

        
        & \textbf{DiSE (ours)} & 0.651 & 0.505 & 0.591 & 0.530 & 0.543 & 0.445 & 0.604 & 0.482 & \textbf{0.544} \\
        \bottomrule
    \end{tabular}
}
\end{table*}

\begin{figure*}[t]
    \centering
    \includegraphics[width=\linewidth]{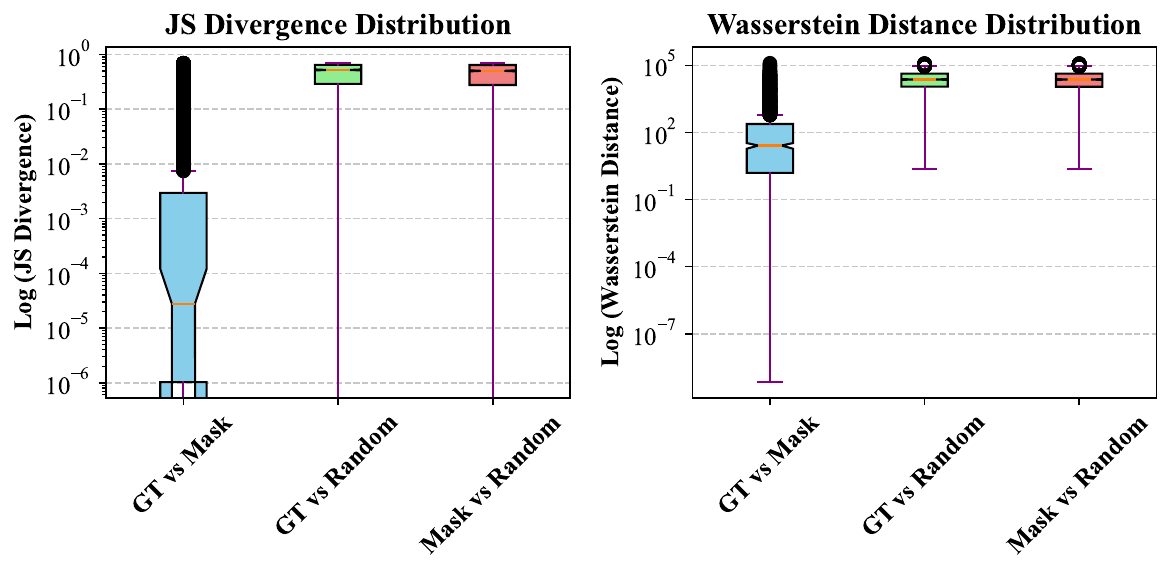}
    \caption{Boxplot of Pairwise Distribution Distances for GT, mask and random tokens using JS Divergence and Wasserstein Distance.}
    \label{fig:distribution-distance-statistic-boxplot}
\end{figure*}

\section{More analyses}
\label{sec:more-analyses}

\subsection{Rank Distribution Statistics for Fixed Sequences and Positions}
\label{sec:rank-distribution-statistics-more}

\begin{figure*}[t]
    \centering
    \includegraphics[width=\linewidth]{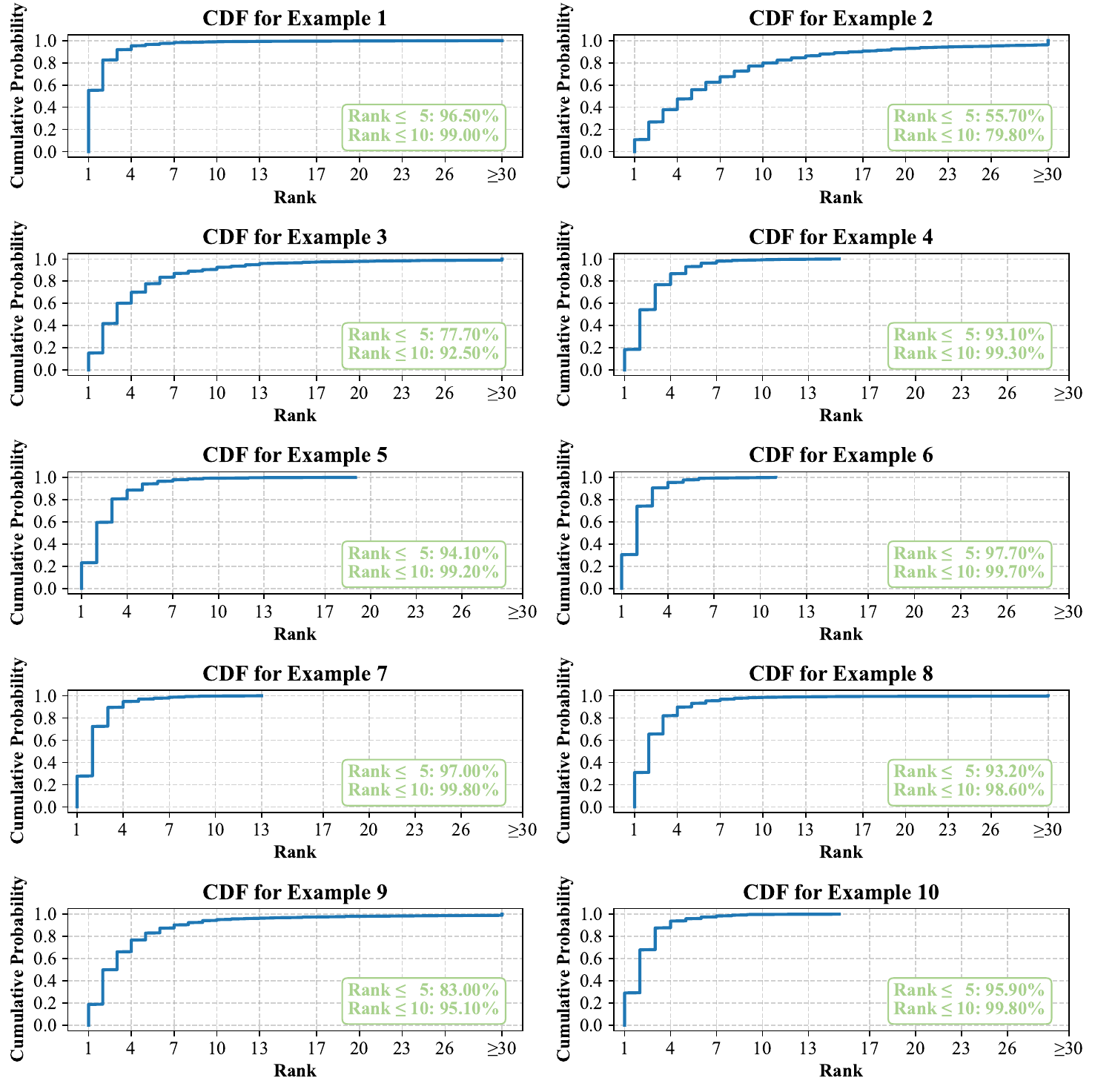}
    \caption{Cumulative distribution function (CDF) of GT token probability ranks for fixed sequence and position. (Part 1)}
    \label{fig:statistic-GT-rank-for-random-token-all-subfigures-part-1}
\end{figure*}

\begin{figure*}[t]
    \centering
    \includegraphics[width=\linewidth]{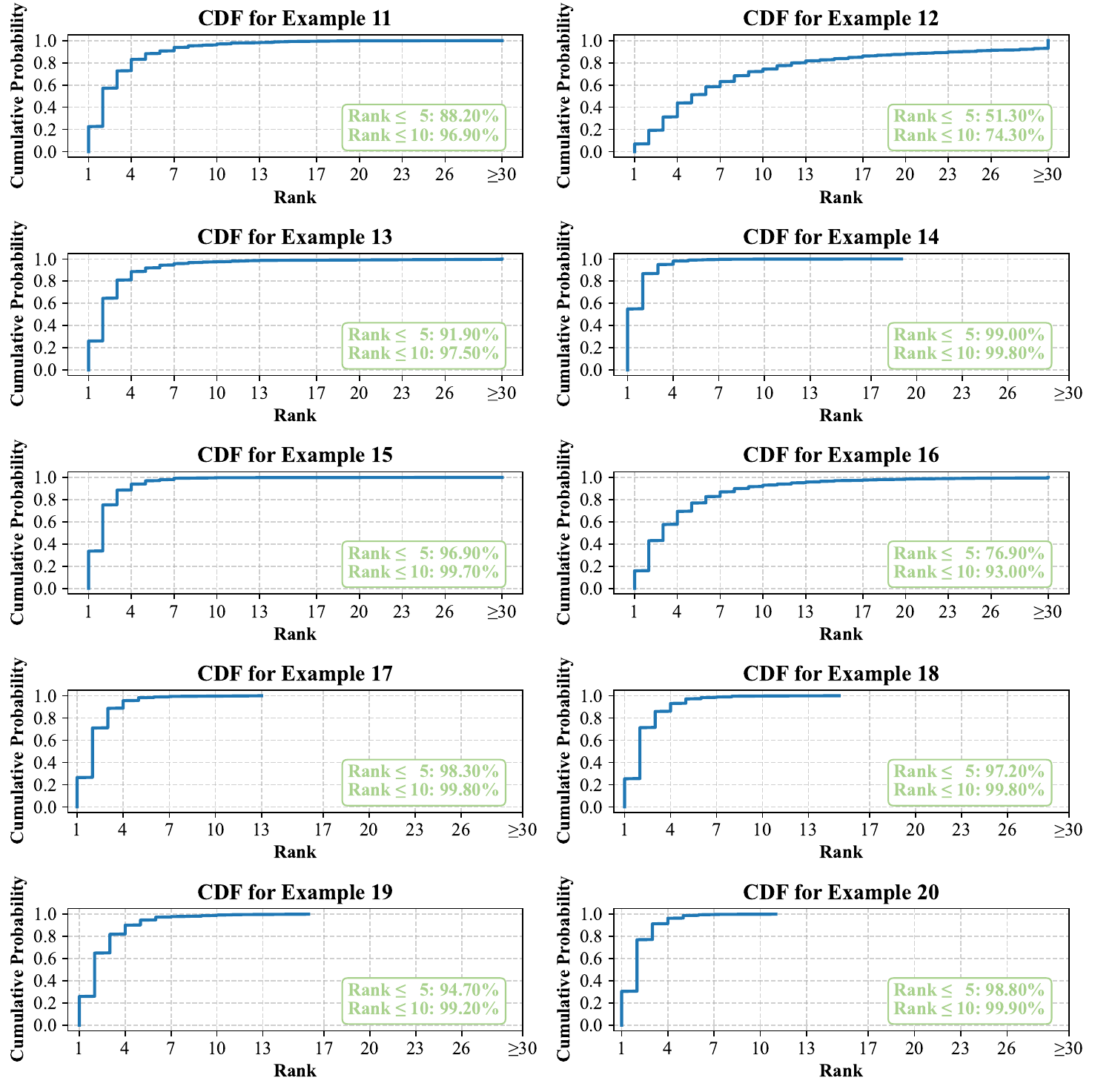}
    \caption{Cumulative distribution function (CDF) of GT token probability ranks for fixed sequence and position. (Part 2)}
    \label{fig:statistic-GT-rank-for-random-token-all-subfigures-part-2}
\end{figure*}

For each fixed sequence, we replace the token at a specific position with a random token and evaluate the rank of the GT token within the predicted probability distribution at that position. For each sequence-position pair, 1,000 random tokens are sampled from the vocabulary and the resulting GT rank distributions are analyzed. As illustrated in Figure~\ref{fig:statistic-GT-rank-for-random-token-all-subfigures-part-1} and Figure~\ref{fig:statistic-GT-rank-for-random-token-all-subfigures-part-2}, we analyzed 20 different sequence-position pairs. The results indicate that the GT token predominantly occupies low ranks within the distribution. Although there is some variation between different sequences and positions, even in the worst-case scenario, the probability that the GT token appears within the top 10 ranks remains above $70\%$. These findings further confirm the generalization ability of the dLLM under random perturbations.

\subsection{Boxplot for Distribution Distances}
\label{sec:boxplot-for-distribution-distances}

For each fixed sequence, we replace the token at a specific position with the GT token, the mask token, and a random token, respectively, and compute the pairwise distribution distances using JS Divergence and the Wasserstein Distance. We analyze a total of 2,509 instances and the resulting boxplot is shown in Figure~\ref{fig:distribution-distance-statistic-boxplot}. The results indicate that the distribution distances between GT and mask tokens are consistently much smaller than those between either of them and random tokens, further demonstrating the effectiveness of the token regeneration approach.

\end{document}